\documentclass[sigconf, nonacm]{acmart}
\AtBeginDocument{%
  }

\copyrightyear{2026}
\acmYear{2026}
\setcopyright{none} 
\acmConference[ICMR '26]{ACM International Conference on Multimedia Retrieval}{June 2026}{TBD} 
\acmBooktitle{Proceedings of the 2026 ACM International Conference on Multimedia Retrieval (ICMR '26), June 2026}
\acmPrice{} 
\acmDOI{}   
\acmISBN{}  

\usepackage{mathtools}
\usepackage{subcaption} 
\usepackage{multirow}   
\usepackage{colortbl}   
\usepackage{booktabs}   
\makeatletter
\newcommand{\boldpara}{%
  \@startsection{paragraph}{4}{0pt}%
  {1.2ex \@plus .3ex \@minus .2ex}%
  {-0.8em}%
  {\bfseries}}
\makeatother
\begin{document}

\title{RHVI-FDD: A Hierarchical Decoupling Framework for Low-Light Image Enhancement}



\author{Junhao Yang}
\affiliation{%
  \institution{Jilin University}
  \city{Changchun}
  \country{China}
}
\email{yjh24@mails.jlu.edu.cn}

\author{Bo Yang}
\affiliation{%
  \institution{Jilin University}
  \city{Changchun}
  \country{China}
}
\email{ybo@jlu.edu.cn}

\author{Hongwei Ge}
\affiliation{%
  \institution{Dalian University of Technology}
  \city{Dalian}
  \country{China}
}
\email{gehw@dlut.edu.cn}

\author{Yanchun Liang}
\affiliation{%
  \institution{Jilin University}
  \city{Changchun}
  \country{China}
}
\email{ycliang@jlu.edu.cn}

\author{Heow Pueh Lee}
\affiliation{%
  \institution{National University of Singapore}
  \country{Singapore}
}
\email{mpeleehp@nus.edu.sg}

\author{Chunguo Wu}
\authornote{Corresponding author.} 
\affiliation{%
  \institution{Jilin University}
  \city{Changchun}
  \country{China}
}
\email{wucg@jlu.edu.cn}

\renewcommand{\shortauthors}{Yang, et al.}

\begin{abstract}
Low-light images often suffer from severe noise, detail loss, and color distortion, which hinder downstream multimedia analysis and retrieval tasks. The degradation in low-light images is complex: luminance and chrominance are coupled, while within the chrominance, noise and details are deeply entangled, preventing existing methods from simultaneously correcting color distortion, suppressing noise, and preserving fine details. To tackle the above challenges, we propose a novel hierarchical decoupling framework (RHVI-FDD). At the macro level, we introduce the RHVI transform, which mitigates the estimation bias caused by input noise and enables robust luminance-chrominance decoupling. At the micro level, we design a Frequency-Domain Decoupling (FDD) module with three branches for further feature separation. Using the Discrete Cosine Transform, we decompose chrominance features into low, mid, and high-frequency bands that predominantly represent global tone, local details, and noise components, which are then processed by tailored expert networks in a divide-and-conquer manner and fused via an adaptive gating module for content-aware fusion. Extensive experiments on multiple low-light datasets demonstrate that our method consistently outperforms existing state-of-the-art approaches in both objective metrics and subjective visual quality.
\end{abstract}

\begin{teaserfigure}
  \centering
  \includegraphics[width=0.95\textwidth]{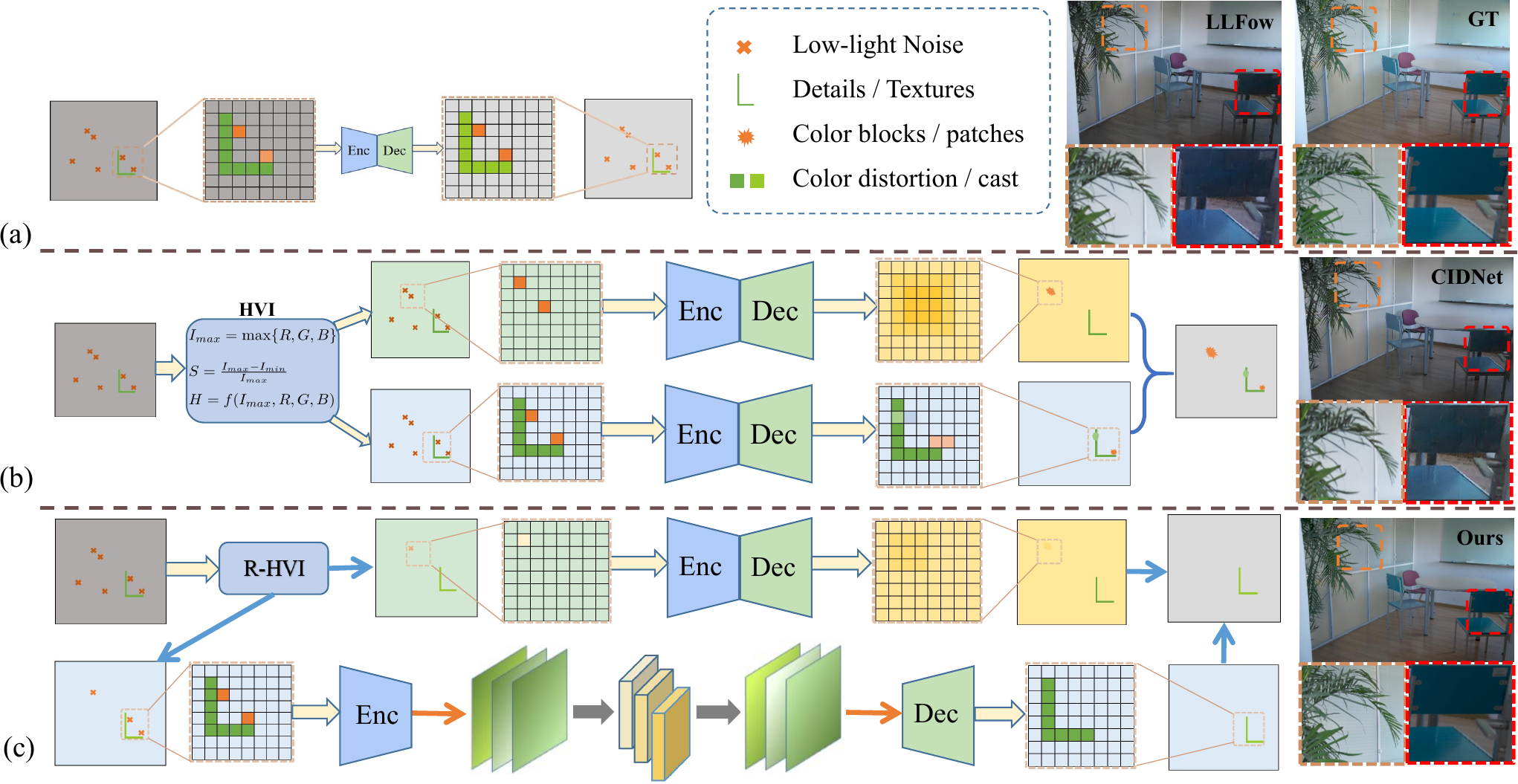}
  \caption{Methods overview and visual comparison. (a) Conventional methods, which often produce color bias and brightness artifacts due to inherent high color sensitivity in sRGB. (b) HVI mitigates color shifts via luminance–chrominance decoupling, but coarse illumination estimation under noise induces overexposure, uneven brightness, and entangled chrominance features. (c) Ours: employs the optimized RHVI transformation for more robust luminance-chrominance decoupling, and applies feature-level decoupling in the chrominance map to separately handle tone, details, and noise.}
  \Description{Comparison of three methods. (a) shows sRGB methods with color artifacts. (b) shows HVI methods with noise issues. (c) shows the proposed RHVI method with clean details and balanced colors.}
  \label{fig:view}
\end{teaserfigure}

\maketitle

\section{Introduction}
\label{sec:intro}

Low-light image enhancement (LLIE) is a fundamental computer vision task addressing a highly ill-posed image restoration problem~\cite{li2021low}. Its intrinsic difficulty stems from the physical formation of low-light images. First, extreme photon flux sparsity results in very low sensor signal-to-noise ratio (SNR)~\cite{kellman2005image}. Second, photoelectric conversion introduces complex Poisson-Gaussian noise with signal-dependent, nonlinear statistics that violate conventional additive white Gaussian noise (AWGN) assumptions~\cite{middleton2002non,pauluzzi2002comparison}. Finally, these signals are nonlinearly mapped into the sRGB space optimized for human perception, deeply entangling luminance and chrominance~\cite{schwartz2018deepisp}. This multi-stage physical degradation implies that any end-to-end network faces extreme challenges~\cite{schwartz2018deepisp}. Therefore, a divide-and-conquer decoupling strategy aligned with the physical formation of low-light images constitutes a reasonable method for handling the above highly entangled degradations. 

The evolution of prior research represents an initial step toward feature decoupling~\cite{land1971lightness}. A key challenge is macro-level entanglement, where luminance and chrominance are inherently coupled in sRGB~\cite{gevers2012color}. Optimized for human perception, sRGB's nonlinear gamma compression and channel mixing intertwine luminance and chrominance, causing noticeable color shifts and saturation drifts under low-light conditions (Fig.~\ref{fig:view}(a)). To address this, alternative color spaces have been developed~\cite{joblove1978color}. HVI~\cite{yan2025hvi} decomposes an input image into a luminance map and a chrominance map, each processed by separate enhancement modules, achieving effective luminance–chrominance decoupling.

Even after macro-level separation, the chrominance component is a highly mixed spatial-domain signal, containing random noise, fine textures, and smooth structure/tone~\cite{wei2021physics,xu2020learning}. For spatial-domain networks, processing such mixed features is inherently a ``zero-sum game'': detail enhancement inevitably amplifies noise, while strong denoising blurs fine structures~\cite{wang2020experiment,yang2023lightingnet,li2015low}. The existing HVI method exhibits this issue: it estimates illumination using the highly noise-sensitive Max-RGB theory, where noise peaks are often mistaken for lighting and preserved or even amplified, degrading the luminance map (Fig.~\ref{fig:view}(b)). The chrominance computation relies on this luminance, entangling noise with true chrominance at the pixel level and embedding both into the feature map, aggravating micro-level entanglement~\cite{yan2025hvi}.

To systematically address the above challenges, we propose a novel hierarchical decoupling framework that decomposes the complex LLIE task into two interrelated sub-tasks handled in a hierarchical manner. The main contributions of our work are as follows:

\begin{itemize}
\item We systematically identify that LLIE suffers from two levels of feature entanglement, the coupling of luminance-chrominance at the macro level and noise-details at the micro level, and propose a novel hierarchical decoupling framework to systematically address LLIE.
\item We reveal the inherent noise sensitivity of the Max-RGB prior in traditional color space transformations. To address this, we introduce the RHVI transform together with a robust illumination refinement module to refine illumination estimation bias at the source.
\item We systematically introduce frequency-domain processing into chrominance feature refinement by designing the FDD framework, which performs divide-and-conquer processing, achieving precise and differentiated handling of noise, details, and tone.
\item Extensive experiments demonstrate state-of-the-art performance. Despite introducing only modest parameters and computational overhead, RHVI-FDD consistently yields performance improvements across different LLIE methods.
\end{itemize}
\setlength{\textfloatsep}{6pt}
\setlength{\floatsep}{6pt}
\setlength{\intextsep}{6pt}
\setlength{\abovecaptionskip}{2pt}
\setlength{\belowcaptionskip}{0pt}
\section{Related Work}
\label{sec:formatting}

\begin{figure*}[t]
    \centering
    \includegraphics[width=\textwidth]{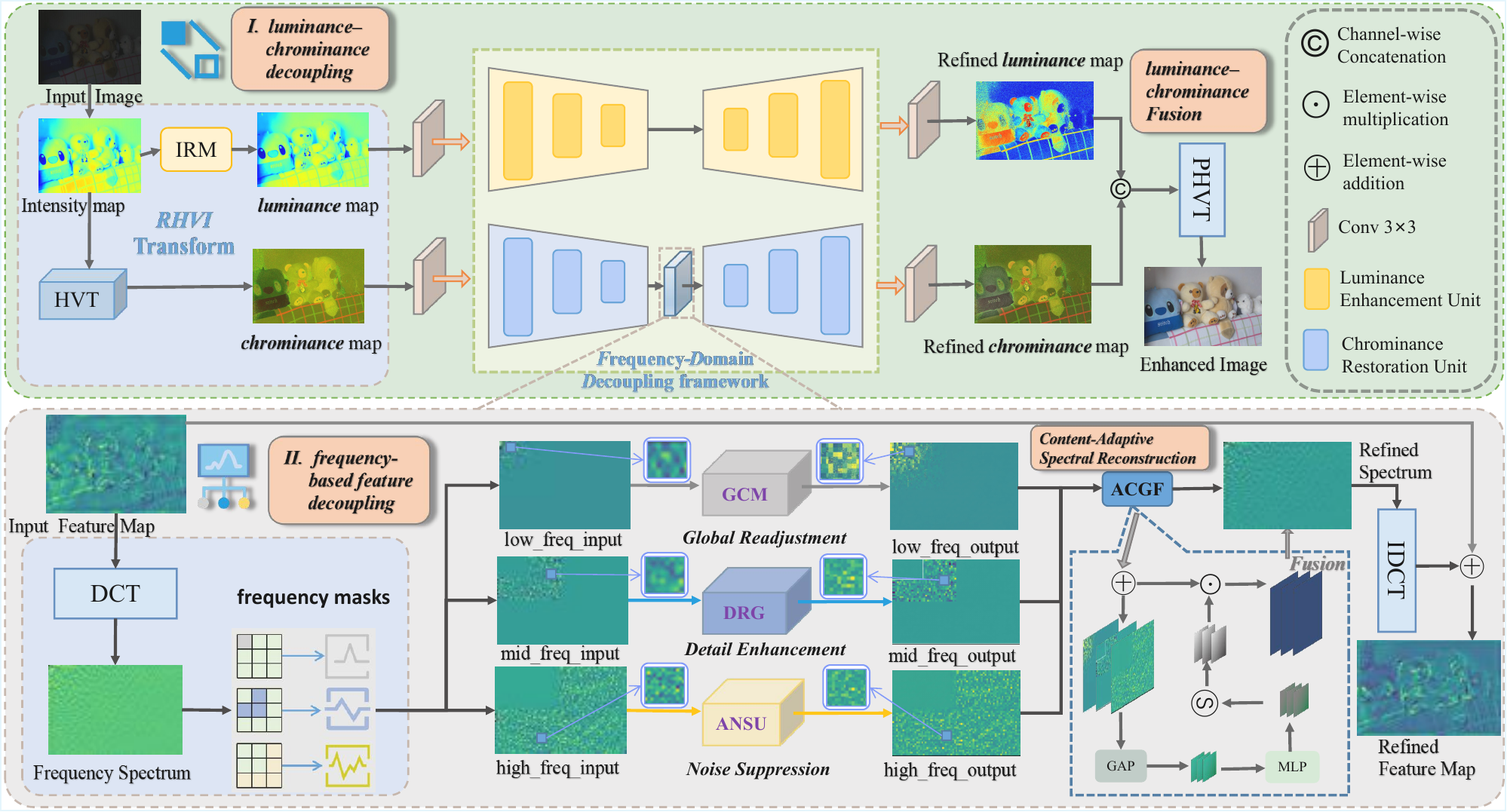}
    \Description{Diagram of the proposed method. Part I shows the RHVI transform process decoupling luminance and chrominance. Part II shows the FDD framework processing features in the frequency domain with three distinct branches for hue, detail, and noise, followed by a fusion module.}
    \caption{(I) RHVI refines noise-corrupted chrominance in HVI via IRM, achieving robust luminance-chrominance decoupling and generating luminance and chrominance maps, which are processed by a dual-branch network for the feature refinement. (II) The FDD framework, embedded in the chrominance branch, projects features into the frequency domain via DCT and splits them into three bands for hue-structure, detail, and noise. Each band is handled by expert networks (GCM, DRG, ANSU) and fused by the fusion module (ACGF).}
    \label{fig:method}
\end{figure*}
\subsection{Deep Learning-Based LLIE}
\setlength{\textfloatsep}{10pt}
\setlength{\floatsep}{10pt}
\setlength{\intextsep}{10pt}
\setlength{\abovecaptionskip}{10pt}
\setlength{\belowcaptionskip}{0pt}
Deep learning-based methods have been widely adopted in LLIE. One major direction builds upon the Retinex-based decoupling idea, as Wei~et~al.~\cite{wei2018deep} proposed RetinexNet, restoring illumination and reflectance components via separate sub-networks to enhance interpretability, and KinD~\cite{zhang2019kindling} and KinD++~\cite{zhang2021beyond} by Zhang~et~al. refined the decomposition and constraint strategies. Another direction directly learns an end-to-end mapping from low-light to normal-light images~\cite{lore2017llnet,gharbi2017deep,zhang2019kindling,cai2018learning}, which progressively enhances luminance preservation and structural fidelity. Later CNN-based architectures, including MIR-Net~\cite{zamir2022learning} and DLN~\cite{wang2020lightening}, further improved enhancement quality. Vision Transformers have advanced LLIE. RetinexFormer by Cai~et~al.~\cite{cai2023retinexformer} and LLFormer by Wang~et~al.~\cite{wang2023ultra} integrate long-range dependencies. GAN-based approaches, such as EnlightenGAN by Jiang~et~al.~\cite{jiang2021enlightengan}, leverage unpaired adversarial learning. Diffusion models, notably Diff-Retinex by Yi~et~al.~\cite{yi2023diff}, demonstrated strong post-enhancement generation. However, most methods operate in the sRGB space, which often causes color shifts and fails to balance denoising and detail preservation~\cite{liu2021benchmarking}.
\subsection{Color Spaces for LLIE}
To address the color shift problem in sRGB, some works explore alternative color spaces, such as HSV~\cite{sural2002segmentation} and YCbCr~\cite{shaik2015comparative}. However, these general-purpose spaces are not designed for low-light enhancement. Some works~\cite{shaik2015comparative,sural2002segmentation} noted that HSV suffers from ``red discontinuity'' and ``black-plane noise''~\cite{yan2025hvi}, while other studies~\cite{yan2025hvi,asmare2009color} found that the Y channel in YCbCr remains partially coupled with CbCr, leading to unnatural color shifts. Yan~et~al.~\cite{yan2025hvi} proposed HVI, a color space specifically designed for low-light enhancement. However, its strong reliance on a noise-sensitive heuristic embeds noise into the chrominance features early, making it harder for subsequent networks to process. 
\subsection{Frequency-Based Image Processing}
Frequency-domain analysis offers a distinct perspective for image processing by exploiting the separability of signal components by frequency. 
In image denoising, methods such as DnCNN~\cite{zhao2018low} and FFDNet~\cite{zhang2018ffdnet} demonstrate CNNs' ability to learn frequency-aware features. Herbreteau~et~al.~\cite{herbreteau2022dct2net} showed that a DCT denoiser can be seen as a shallow CNN, and Karaoglu~et~al.~\cite{karaoglu2023dctnet} developed DCTNet for grayscale and color image denoising. 
Recently, frequency-domain features have been explored for LLIE. The FourLLIE model by Wang~et~al.~\cite{wang2023fourllie} analyzes amplitude spectra for illumination consistency, 
while the FreqLLIE method by He~et~al.~\cite{he2023low} jointly exploits amplitude and phase to enhance detail and contrast. However, most methods treat frequency information as auxiliary or apply it only in limited stages. Introducing a comprehensive frequency-domain framework for LLIE remains an unexplored and highly promising direction.
\section{Methodology}
We propose a hierarchical decoupling framework to systematically address LLIE, as shown in Fig.~\ref{fig:method}. First, to overcome the noise sensitivity of HVI in luminance-chrominance decoupling (Sec.~\ref{sec:noise_sensitivity}), we introduce the RHVI transform for more robust separation (Sec.~\ref{sec:rhvi}). Second, to tackle feature entanglement in chrominance maps (Sec.~\ref{sec:entanglement}), we design a frequency-based decoupling framework that processes features in a divide-and-conquer manner (Sec.~\ref{sec:fdd}).
\subsection{Noise Sensitivity in HVI}
\label{sec:noise_sensitivity}
The HVI color space was recently proposed to address the color distortion of sRGB by decoupling the image into a luminance component ($I$) and two chrominance components (${H}$, ${V}$). The entire HVI pipeline is fundamentally constructed upon an initial luminance estimation, which is defined by the Max-RGB rule:\begin{equation}
I_{\max}(x) = \max_{c \in \{R,G,B\}} I_c(x).
\end{equation}
However, this design, while enabling decoupling, relies on a clean signal prior and overlooks the heavy noise inherent in low-light images. Under low SNR conditions, the direct application of the max operator renders the estimation highly susceptible to input sensor noise. To expose its intrinsic limitation, we model the formation process of low-light images. Under insufficient illumination, an observed image can be expressed as the sum of a clean underlying signal and a complex noise distribution:
\begin{equation}
I_{\text{low}}(x) = S_{\text{clean}}(x) + N(x),
\end{equation}
where $x$ denotes the pixel location and $N(x)$ aggregates multiple noise sources---most notably shot noise and readout noise---that often manifest as random, isolated bright spikes~\cite{wei2021physics}. The illumination map estimated by the HVI transform is therefore noise-contaminated and computed as
\begin{equation}
I_{\text{max\_noisy}}(x) = \max_{c \in \{R,G,B\}} (S_{\text{clean},c}(x) + N_c(x)).
\end{equation}
\setlength{\textfloatsep}{6pt}
\setlength{\floatsep}{6pt}
\setlength{\intextsep}{6pt}
\setlength{\abovecaptionskip}{2pt}
\setlength{\belowcaptionskip}{2pt}
\begin{figure}[t]
\centering
\includegraphics[width=\linewidth]{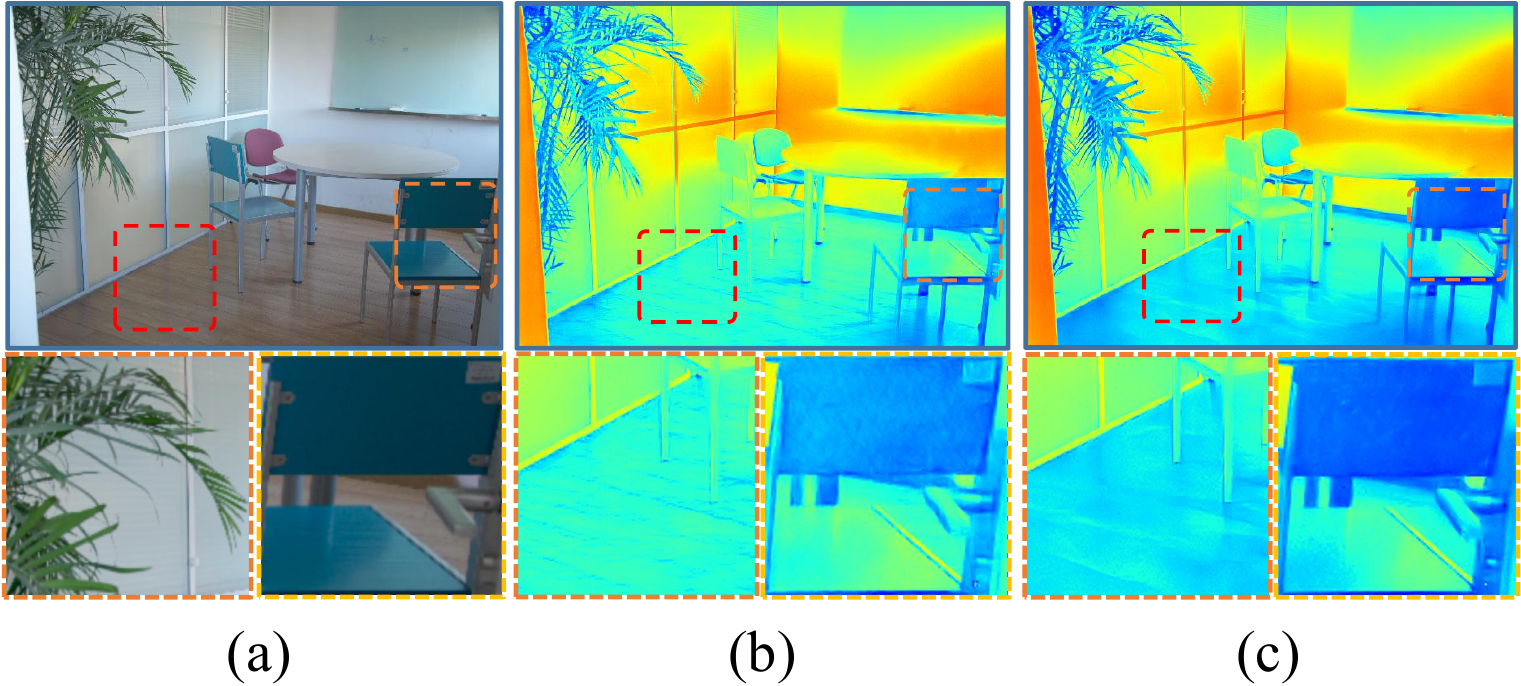}
\caption{Comparison of luminance maps.
(a) Normal-light image. (b) Luminance map produced by the luminance branch of HVI. (c) Luminance map obtained by feeding the RHVI-transformed result into the same branch.}
\label{fig:luminance_comparison}
\end{figure}
\setlength{\textfloatsep}{10pt}
\setlength{\floatsep}{10pt}
\setlength{\intextsep}{10pt}
\setlength{\abovecaptionskip}{10pt}
\setlength{\belowcaptionskip}{0pt}

A single positive noise spike in any channel can dominate the max operator. To quantify this effect, we define the ideal, noise-free illumination map $I_{\text{max\_clean}}$, which faithfully represents the underlying scene structure: 
\begin{equation}
I_{\text{max\_clean}}(x) = \max_{c \in \{R,G,B\}} (S_{\text{clean},c}(x)).
\end{equation}
Their difference forms the noise-induced bias term:
\begin{align}
\delta_N(x) &= I_{\text{max\_noisy}}(x) - I_{\text{max\_clean}}(x) \nonumber \\
&= \max_{\mathclap{c \in \{R,G,B\}}} (S_{\text{clean},c}(x) + N_c(x)) - \max_{\mathclap{c \in \{R,G,B\}}} (S_{\text{clean},c}(x)).
\end{align}
Because the max operator is extremely sensitive to strong positive spikes, $\delta_N(x)$ is statistically non-negative and causes HVI to consistently overestimate illumination. When a network directly enhances the contaminated illumination map, amplified noise in smooth dark regions (chair in Fig.~\ref{fig:luminance_comparison}) tends to be reconstructed as color blotches or artifacts~\cite{zhang2021beyond}, while in extremely dark, noise-dense regions (floor in Fig.~\ref{fig:luminance_comparison}) aggressive enhancement produces local overexposure. Moreover, and more critically for multimedia tasks, the noise-induced variance disrupts the structural continuity of the illumination component. This structural corruption severely hinders downstream feature extraction algorithms~\cite{salahat2017recent,lew2006content}, which rely on stable gradients for accurate content retrieval.
\subsection{RHVI Transform}
\setlength{\textfloatsep}{6pt}
\setlength{\floatsep}{6pt}
\setlength{\intextsep}{6pt}
\setlength{\abovecaptionskip}{2pt}
\setlength{\belowcaptionskip}{2pt}
\begin{figure}[t]
    \centering
    \includegraphics[width=\linewidth]{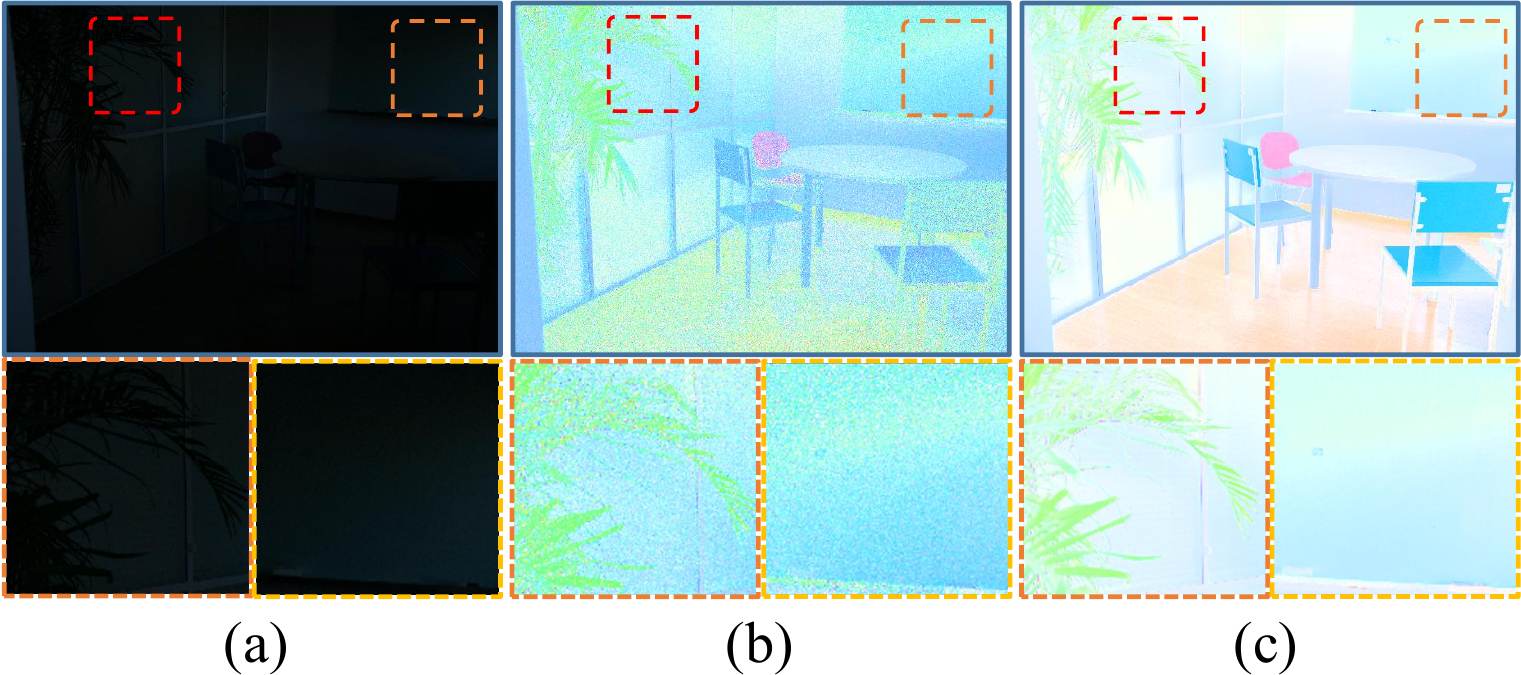}
    \caption{Comparison of chrominance maps.
(a) Low-light image. (b) Chrominance map of (a) after HVI decoupling. (c) Chrominance map of the ground-truth image processed similarly. Both (b) and (c) are visualized using the same pseudocolor reconstruction.}
    \label{fig:chrominance_maps}
\end{figure}
\setlength{\textfloatsep}{10pt}
\setlength{\floatsep}{10pt}
\setlength{\intextsep}{10pt}
\setlength{\abovecaptionskip}{10pt}
\setlength{\belowcaptionskip}{0pt}
\label{sec:rhvi}
To address the noise sensitivity of the HVI, we propose a robust luminance--chrominance decoupling method, the RHVI transformation, which leverages a lightweight Illumination Refinement Module (IRM) to produce a cleaner illumination map.

As illustrated in Fig.~\ref{fig:modules}, the $I_{\text{initial}}$ as input first passes through a $1\times1$ convolution and then a CoreBlock comprising a $5\times5$ depthwise separable convolution, GELU activation, and batch normalization (BN). The $5\times5$ kernel provides a local receptive field to smooth the illumination map and aggregate neighboring information. A residual connection adds the CoreBlock output to the $1\times1$-convolved $I_{\text{initial}}$, and a final $1\times1$ convolution compresses channels to produce the refined illumination map $I_{\text{refined}}$.
The illumination map generated by the RHVI transformer is smoother and more consistent with the actual illumination distribution (Fig.~\ref{fig:luminance_comparison}). It mitigates overexposure and color blotch artifacts, achieving more accurate luminance-chrominance decoupling.
\subsection{Feature Entanglement in Chrominance Map}
\label{sec:entanglement}
The chrominance components (${H}$, ${V}$)  isolate a classic challenge in low-light enhancement: restoring the highly degraded chrominance map, which is a core bottleneck that also plagues other mainstream LLIE methods using lightness-chrominance separation, such as Retinex-based~\cite{yang2021sparse,li2018structure} and color-space-based methods. Under dim illumination, photon scarcity causes severe signal-to-noise ratio (SNR) drops in color channels, leading to saturation attenuation and color distortion~\cite{fan2022multiscale}. More critically, as shown in Fig.~\ref{fig:chrominance_maps}, these noise components become strongly entangled with semantically meaningful local texture details in the spatial domain. We term this challenge microscopic feature entanglement.

To formalize this problem, we model the deep features extracted from this degraded chrominance map:
\begin{equation}
F_{\text{\!spatial\!}} = \mathcal{\!T}(F_{\text{\!clean\!}}, F_{\text{\!noise\!}}) = \mathcal{\!T}(f(F_{\text{\!structure\!}}, F_{\text{\!detail\!}}), F_{\text{\!noise\!}}),
\label{eq:7} 
\end{equation}
where $F_{\text{noise}}$ represents globally distributed noise perturbations, while $F_{\text{clean}}$ denotes the ideal noise-free features composed of the structural and tonal component $F_{\text{structure}}$ and the fine-texture component $F_{\text{detail}}$. The function $f(\cdot)$ characterizes the spatial dependency between $F_{\text{structure}}$ and $F_{\text{detail}}$, and $\mathcal{T}(\cdot)$ models the complex, implicit entanglement induced by degradation and feature extraction. Traditional approaches attempt to learn an equally complex inverse mapping $\mathcal{T}^{-1}(\cdot)$, often face a ``zero-sum game'' between detail fidelity and noise suppression.

\subsection{Frequency-Domain Decoupling}
\label{sec:fdd}
We turn to the frequency domain, which offers a more structured representation for signal separation. As illustrated in Fig.~\ref{fig:method}, we therefore introduce a Frequency-domain Decoupling (FDD) framework that resolves this complex entanglement by transforming it into a more separable frequency-domain problem. 
\setlength{\textfloatsep}{6pt}
\setlength{\floatsep}{6pt}
\setlength{\intextsep}{6pt}
\setlength{\abovecaptionskip}{3pt}
\setlength{\belowcaptionskip}{2pt}
\begin{figure*}[t]
    \centering
    \includegraphics[width=\textwidth]{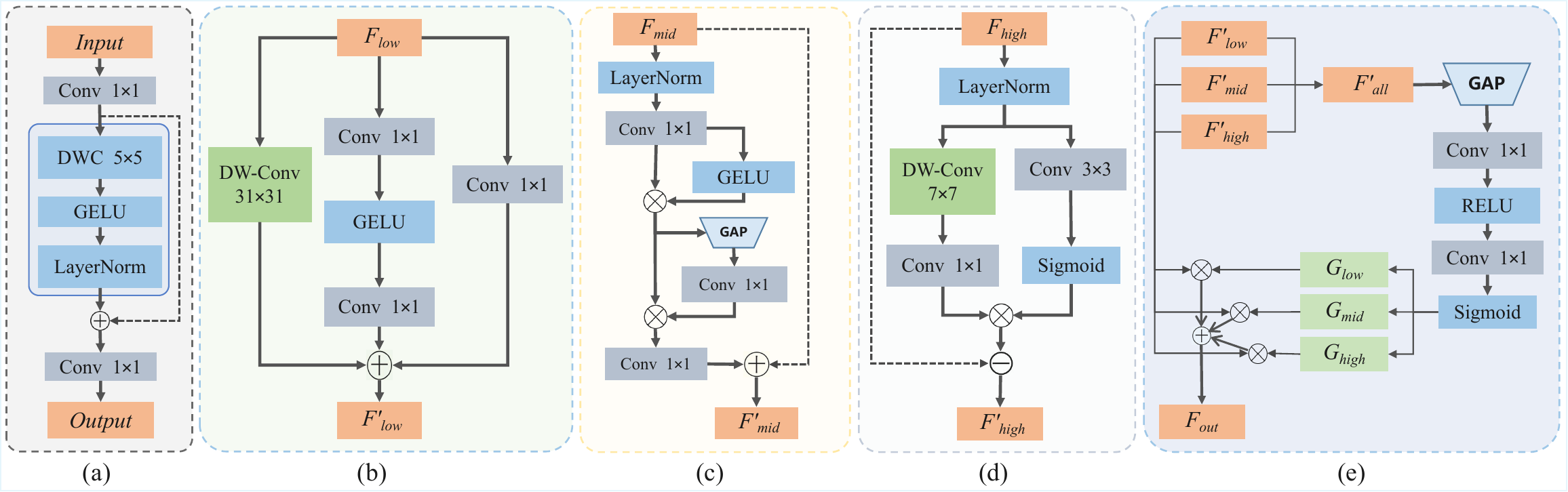}
    \Description{Schematic diagrams of the five proposed modules. (a) shows the Illumination Refinement Module. (b) shows the Global Context Modulator. (c) shows the Detail Refinement Gated Block. (d) shows the Adaptive Noise Suppression Unit. (e) shows the Adaptive Channel-Guided Fusion module.}
    \caption{Detailed architectures of the proposed modules: (a) Illumination Refinement Module (IRM), (b) Global Context Modulator (GCM), (c) Detail Refinement Gated Block (DRG), (d) Adaptive Noise Suppression Unit (ANSU), and (e) Adaptive Channel-Guided Fusion (ACGF).}
    \label{fig:modules}
\end{figure*}
\setlength{\textfloatsep}{10pt}
\setlength{\floatsep}{10pt}
\setlength{\intextsep}{10pt}
\setlength{\abovecaptionskip}{10pt}
\setlength{\belowcaptionskip}{0pt}
\subsubsection{Feature Decoupling via DCT Transform}
\begin{figure}[t]
    \centering
    \begin{subfigure}{\linewidth}
        \centering
        \includegraphics[width=0.32\linewidth]{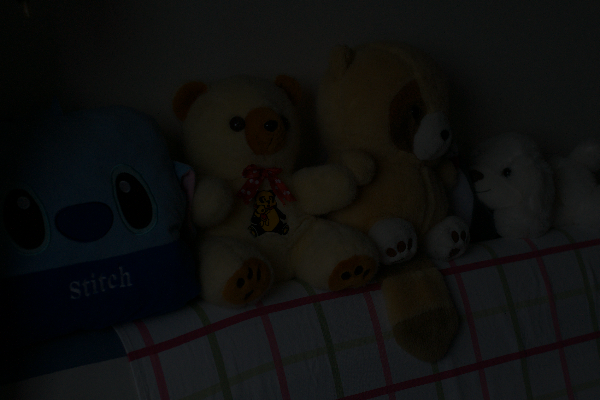}
        \includegraphics[width=0.32\linewidth]{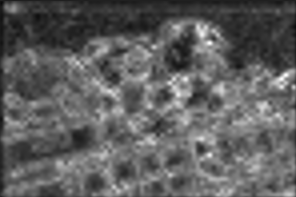}
        \includegraphics[width=0.32\linewidth]{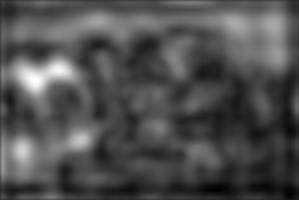}
    \end{subfigure}

    \vspace{-3pt}
    \begin{subfigure}{\linewidth}
        \centering
        \footnotesize
        \parbox{0.32\linewidth}{\centering (a) Low-light Input}%
        \parbox{0.32\linewidth}{\centering (b) Feature Map}%
        \parbox{0.32\linewidth}{\centering (c) Low-band Response}%
    \end{subfigure}

    \vspace{4pt}
    \begin{subfigure}{\linewidth}
        \centering
        \includegraphics[width=0.32\linewidth]{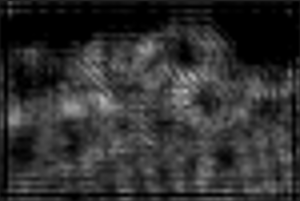}
        \includegraphics[width=0.32\linewidth]{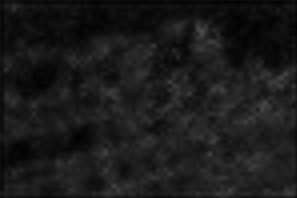}
        \includegraphics[width=0.32\linewidth]{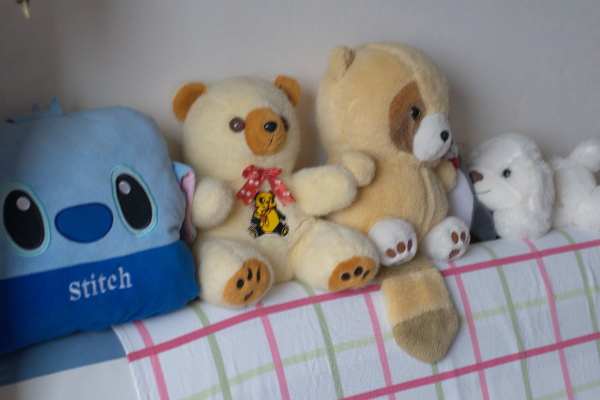}
    \end{subfigure}

    \vspace{-3pt}
    \begin{subfigure}{\linewidth}
        \centering
        \footnotesize
        \parbox{0.32\linewidth}{\centering (d) Mid-band Response}%
        \parbox{0.32\linewidth}{\centering (e) High-band Response}%
        \parbox{0.32\linewidth}{\centering (f) Reference GT}%
    \end{subfigure}

    \caption{Spatial reconstructions of downsampled bottleneck features, obtained by projecting individual frequency bands back to the spatial domain via IDCT, revealing distinct frequency-band responses and supporting the underlying frequency decoupling assumption.}
    \label{fig:spectral_analysis}
\end{figure}

\mbox{}\\
FDD's first step is to project the spatial features $F_{\text{spatial}}$ into the frequency domain $F_{\text{freq}}$. We adopt the two-dimensional Discrete Cosine Transform (2D-DCT) for its well-known energy compaction property~\cite{wang2023discrete}. To provide a stable structural prior under low-SNR conditions—where learnable or adaptive frequency boundaries can be easily driven by noise-dominated spectra and become unstable—we adopt a fixed-ratio partition strategy. Specifically, based on the energy compaction property of DCT~\cite{wang2023discrete}, we define binary masks $\mathcal{M}=\{M_{\text{low}},M_{\text{mid}},M_{\text{high}}\}$ to partition the spectrum:
\begin{equation}
M_{\text{low}}(i, j) = \begin{cases} \!1, & \text{if } 0 \le i < \alpha H \text{ and } 0 \le j < \alpha W \\ \!0, & \text{otherwise} \end{cases},
\end{equation}
\begin{equation}
M_{\text{high}}(i, j) = \begin{cases} 1, & \text{if } i \ge \beta H \text{ or } j \ge \beta W \\ 0, & \text{otherwise} \end{cases},
\end{equation}
\begin{equation}
M_{\text{mid}} = \mathbf{1} - M_{\text{low}} - M_{\text{high}},
\end{equation}
where $\mathbf{1}$ denotes an all-one matrix. Here, $\alpha=1/4$ and $\beta=1/2$ are chosen according to the characteristic energy compaction of natural images in the DCT domain and the statistical behavior of low-light imaging, serving as a stable geometric prior rather than a finely tuned boundary. Importantly, this partition is not intended to enforce a strict semantic separation, but rather to provide a coarse and stable decomposition that facilitates divide-and-conquer learning. The final restoration quality is therefore not determined by the exact frequency boundaries, but by the complementary behaviors of the band-specific expert networks and the subsequent adaptive fusion mechanism. Finally, the individual frequency components $F_k$ are obtained by applying the masks $M_k$ to the full spectrum $F_{\text{freq}}$ via element-wise multiplication:
\begin{equation}
F_{k} = F_{\text{freq}} \odot M_{k}, \quad k \in \{\text{low, mid, high}\}.
\end{equation}
Applying these masks yields three components characterized by statistical dominance rather than absolute physical separation. $F_{\text{low}}$ is dominated by global structure and color tone; $F_{\text{mid}}$ encapsulates the majority of semantic textures; and $F_{\text{high}}$, while containing fine edges, is statistically dominated by random noise spikes~\cite{xu2022structure,zhou2023low,liang2023low}. As shown in Fig.~\ref{fig:spectral_analysis}, this frequency prior decomposes the spatially entangled features into independent component sub-bands.

\subsubsection{Expert Networks for Sub-Band Processing}
\mbox{}\\
After frequency decomposition, we focus on targeted processing for each sub-band. The DCT spectrum is not an unstructured numerical matrix but exhibits intrinsic statistical regularities and characteristic energy distributions~\cite{xu2021dct}, indicating that different frequency components convey distinct and learnable patterns. Prior studies~\cite{xu2020learning,chen2018learning,zhang2018ffdnet} have demonstrated that CNNs can effectively model such structured frequency information. Motivated by these insights, we design three CNN-based expert networks, each specifically crafted to resolve the unique signal-mixture pattern within its sub-band using tailored structure, thereby handling the residual entanglement that simple filtering cannot address:
\boldpara{Global Context Modulator (GCM).}
The low-frequency component $F_{\text{low}}$ carries the macroscopic structure and overall tonal information but suffers from energy decay and spectral chaos. To address this, GCM decomposes the restoration into parallel branches. First, for Spectral Structure Regularization, we employ a large-kernel depth-wise convolution. Its extensive receptive field captures the long-range correlations of the frequency neighborhood to suppress spectral disorder. Second, for Non-linear Magnitude Re-calibration, a lightweight MLP is used to recover the color energy distribution. To prevent potential over-smoothing, a parallel $1\times1$ residual branch is added. The final refinement is obtained by fusing these branches:
\begin{equation}
F_{\text{low}}' = F_{\text{low}} + F_{\text{DW}}(F_{\text{low}}) + F_{\text{PW}}(F_{\text{low}}) + F_{\text{Res}}(F_{\text{low}}),
\end{equation}
where $F_{\text{DW}}$, $F_{\text{PW}}$, and $F_{\text{Res}}$ denote the mapping functions of the structure, calibration, and residual branches, respectively.
\setlength{\textfloatsep}{6pt}
\setlength{\floatsep}{6pt}
\setlength{\intextsep}{6pt}
\setlength{\abovecaptionskip}{6pt}
\setlength{\belowcaptionskip}{4pt}
\setlength{\textfloatsep}{10pt}
\setlength{\floatsep}{10pt}
\setlength{\intextsep}{10pt}
\setlength{\abovecaptionskip}{10pt}
\setlength{\belowcaptionskip}{0pt}
\boldpara{Detail Refinement Gated Block (DRG-Block).}
This module processes the mid-frequency band, which contains valuable ``spectral fingerprints'' (details) and interfering patterns (noise)~\cite{jattke2022blacksmith}. It is designed for feature selectivity, employing a Gated Mechanism defined by the split-interaction $F_g = F_1 \odot \sigma(F_2)$, where $F_1$ and $F_2$ are split features, with GELU activation $\sigma$ to adaptively amplify valid spectral patterns while suppressing unwanted ones. Subsequently, the gated feature is refined by a Simplified Channel Attention (SCA) module and fused through the residual connection:
\begin{equation}
F'_{\text{mid}} = F_{\text{mid}} + \text{Conv}_{1\times1}(\mathbf{F}_g \odot \phi(W_c \, \text{GAP}(\mathbf{F}_g))),
\end{equation}
where $\phi$ denotes the Sigmoid function and GAP represents global average pooling.

\boldpara{Adaptive Noise Suppression Unit (ANSU).}
This module handles the challenging high-frequency band, a mix of diffuse noise and sparse edge signals. It follows a ``Predict-Regulate-Subtract'' philosophy: a Noise Prediction Branch ($\mathcal{B}_{\text{noise}}$), consisting of a $7 \times 7$ large-kernel depthwise convolution followed by a $1 \times 1$ pointwise convolution, models the statistical patterns of the diffuse noise, while an Adaptive Gating Branch ($\mathcal{B}_{\text{gate}}$) employing a $3 \times 3$ convolution generates a modulation map to protect edge signals. The modulated noise is then adaptively subtracted via
\begin{equation}
F_{\text{high}}' = F_{\text{high}} - \mathcal{B}_{\text{noise}}(\tilde{F}_{\text{high}}) \odot \sigma(\mathcal{B}_{\text{gate}}(\tilde{F}_{\text{high}})),
\end{equation}
where $\sigma$ denotes the Sigmoid function and $\tilde{F}_{\text{high}}$ represents the normalized input.

\subsubsection{Frequency Fusion and Spatial Reconstruction}
\mbox{}\\
After processing by the expert networks, the refined spectral components ($F'_{\text{low}}, F'_{\text{mid}}, F'_{\text{high}}$) must be fused. Although all processing occurs in the frequency domain, the ultimate objective of fusion is to serve spatial-domain visual quality. An ideal mechanism must anticipate how these frequency components will jointly act upon the final image's spatial structure, detail, and color after the IDCT. Simple, fixed fusion rules are insufficient for this task. To this end, we design the Adaptive Channel-Gated Fusion (ACGF) module, which adaptively coordinates frequency components through a globally-aware and dynamically-weighted mechanism~\cite{niaz2024spatially}. 

As shown in Fig.~\ref{fig:modules}, ACGF first concatenates the inputs to extract a global spectral context vector $\mathbf{c}$ via Global Average Pooling. This critical cross-band context $\mathbf{c}$, which summarizes the post-processing global statistics, is fed into a lightweight generator $\mathcal{F}_{\text{gate}}$ to produce dynamic gating weights $\mathbf{g}$. These weights, scaled to the (0, 2) range, adaptively amplify or suppress contributions:
\begin{equation}
g = 2 \cdot \sigma(\mathcal{F}_{\text{gate}}(\mathbf{c})),\quad [\mathbf{g}_{\text{low}},\mathbf{g}_{\text{mid}}, \mathbf{g}_{\text{high}}] = \text{Split}(g).
\end{equation}
The final fusion is achieved via channel-wise weighted summation:
\begin{equation}
F_{\text{fused}} = (F'_{\text{low}} \odot \mathbf{g}_{\text{low}}) + (F'_{\text{mid}} \odot \mathbf{g}_{\text{mid}}) + (F'_{\text{high}} \odot \mathbf{g}_{\text{high}}).
\end{equation}
The resulting $F_{\text{fused}}$ is then transformed back to the spatial domain via IDCT and added to the original feature map in a residual manner, completing the FDD refinement.

\subsection{Loss Function}
Our overall training objective $L_{\text{total}}$ is a composite function that jointly optimizes a main reconstruction loss $L_{\text{main}}$ and an auxiliary illumination loss $L_{\text{aux}}$:
\begin{equation}
L_{\text{total}} = L_{\text{main}} + \lambda_{\text{aux}} L_{\text{aux}}.
\end{equation}
The main loss~\cite{yan2025hvi} ensures the final image quality via   supervision in both the sRGB and HVI color spaces:
\begin{equation}
L_{\text{main}} = L_{\text{sRGB}} + \lambda_{\text{hvi}} L_{\text{HVI}}.
\end{equation}
Both $L_{\text{sRGB}}$ and $L_{\text{HVI}}$ (collectively denoted as $L_{\text{space}}$) are a composite of L1 ($L_1$), perceptual~\cite{rad2019srobb} ($L_{\text{per}}$), edge~\cite{seif2018edge} ($L_{\text{edge}}$), and SSIM~\cite{yan2025hvi} loss ($L_{\text{ssim}}$) terms:
\begin{equation}
L_{\text{space}} = \lambda_1 L_1 + \lambda_p L_{\text{per}} + \lambda_e L_{\text{edge}} + \lambda_s L_{\text{ssim}}.
\end{equation}
To stabilize the IRM and prevent upstream error propagation, we apply an auxiliary $L_{1}$ loss. Unlike $L_2$ loss~\cite{lewkowycz2020training}, which is dominated by outliers, $L_1$ encourages the IRM to recover the underlying structural illumination while suppressing noise. Consistent with the ideal structure defined in Eq. (4), we employ the target's Max-RGB as the noise-free ground truth($I^{\text{GT}}$) to guide the IRM in recovering the physical illumination distribution:
\begin{equation}
L_{\text{aux}} = \|I' - I^{\text{GT}}\|_1.
\end{equation}

\section{Experiments}
\subsection{Experimental Setup}
\boldpara{Datasets.} We evaluate our framework on several widely-used LLIE benchmarks. For paired evaluation, we use the LOL dataset series, including LOLv1~\cite{wei2018deep}, LOLv2-Real and LOLv2-Synthetic~\cite{yang2021sparse}. We also include the SICE~\cite{cai2018learning} and the challenging Sony-Total-Dark dataset, which is a customized version of a SID subset~\cite{chen2018learning}. To assess generalization, we use five unpaired real-world datasets: DICM~\cite{lee2013contrast}, LIME~\cite{guo2016lime}, MEF~\cite{ma2015perceptual}, NPE~\cite{wang2013naturalness}, and VV~\cite{vonikakis2018evaluation}.

\boldpara{Evaluation Metrics.} For all paired datasets, we adopt Peak Signal-to-Noise Ratio (PSNR) and Structural Similarity Index (SSIM)~\cite{wang2004image} as the main metrics, while the Learned Perceptual Image Patch Similarity (LPIPS)~\cite{zhang2018unreasonable} is additionally employed to assess perceptual quality on several datasets. For unpaired datasets, we use the no-reference metric Naturalness Image Quality Evaluator (NIQE)~\cite{mittal2012making}.

\boldpara{Implementation Details.} To fairly validate the effectiveness of our framework, we select a strong dual-branch network: HVI-CIDNet~\cite{yan2025hvi} as our baseline backbone. Our implementation is based on its official code, with two core modifications: we replace its original noise-sensitive HVI with our RHVI transform and embed our FDD into its chrominance branch.
All experiments are conducted in PyTorch 2.7.1 on a single NVIDIA RTX 3090 24 GB GPU. We adopt the Adam optimizer with momentum parameters $\beta_1=0.9$ and $\beta_2=0.999$, set the initial learning rate to $1\times10^{-4}$, and employ cosine annealing for smooth decay. Random horizontal flipping is used for data augmentation. The training of LOLv1 and LOLv2-Real lasts 1500 epochs on $400\times400$ random crops with a batch size of 4, while LOLv2-Synthetic is trained for 500 epochs on full-resolution images with a batch size of 1. For SID, we follow prior work~\cite{yan2025hvi} by converting RAW data to sRGB without gamma correction and train for 1000 epochs on $256\times256$ patches with a batch size of 4. The SICE dataset is trained for 1000 epochs on $160\times160$ patches with a batch size of 10.

\subsection{Quantitative Analysis}
As shown in Table~\ref{tab:quant_final}, we conduct extensive quantitative comparisons with SOTA methods on the LOL datasets. The results demonstrate that our model achieves SOTA or highly competitive performance. Specifically, our method surpasses all approaches in PSNR on all three datasets, achieving 24.822~dB in PSNR and 0.882 in SSIM on the challenging LOLv2-Real dataset and achieves a remarkably low LPIPS score of 0.068 on the LOLv1 dataset.

\setlength{\textfloatsep}{6pt}
\setlength{\floatsep}{6pt}
\setlength{\intextsep}{6pt}
\setlength{\abovecaptionskip}{2pt}
\setlength{\belowcaptionskip}{0pt}
\begin{table*}[t]
\centering
\caption{Quantitative results of PSNR$\uparrow$, SSIM$\uparrow$, and LPIPS$\downarrow$ on the LOL (v1 and v2) datasets. The FLOPs is tested on a single $256 \times 256$ image. The best results are marked in \textbf{\textcolor{red}{red}} and the second-best results are marked in
\textit{\textcolor{cyan}{cyan}} respectively.
 Please note that we apply the GT mean method only on the LOLv1 to minimize errors due to its limited size. }
\label{tab:quant_final}
\setlength{\tabcolsep}{5pt}
\small
\resizebox{\textwidth}{!}{%
\begin{tabular}{@{\hskip0pt} c | c | c c c | c c c | c c c | c c @{\hskip0pt}}
\specialrule{1.2pt}{0pt}{0pt}
\multirow{2}{*}{\centering\arraybackslash\textbf{Methods}} &
\multirow{2}{*}{\centering\arraybackslash\textbf{Reference}} &
\multicolumn{3}{c|}{\textbf{LOLv1}} &
\multicolumn{3}{c|}{\textbf{LOLv2-Real}} &
\multicolumn{3}{c|}{\textbf{LOLv2-Synthetic}} &
\multicolumn{2}{c}{\textbf{Complexity}} \\
\cline{3-13}
 & & PSNR$\uparrow$ & SSIM$\uparrow$ & LPIPS$\downarrow$
 & PSNR$\uparrow$ & SSIM$\uparrow$ & LPIPS$\downarrow$
 & PSNR$\uparrow$ & SSIM$\uparrow$ & LPIPS$\downarrow$
 & Params/M  & FLOPs/G \\
\hline
RetinexNet \cite{wei2018deep} & BMVC'18 & 18.915 & 0.427 & 0.470 & 16.097 & 0.401 & 0.543 & 17.137 & 0.762 & 0.255 & 0.84 & 584.47 \\
KinD \cite{zhang2019kindling} & MM’19 & 23.018 & 0.843 & 0.156 & 17.544 & 0.669 & 0.375 & 18.320 & 0.796 & 0.252 & 8.02 & 34.99 \\
MIRNet \cite{zamir2020learning} & ECCV’20 & 24.140 & 0.842 & 0.131 & 20.357 & 0.782 & 0.317 & 21.927 & 0.833 & 0.209 & 31.76 & 785 \\
RUAS \cite{liu2021retinex} & CVPR'21 & 18.654 & 0.518 & 0.270 & 15.326 & 0.488 & 0.310 & 13.765 & 0.638 & 0.305 & {{0.003}} & {{0.83}} \\
LLFlow \cite{wang2022low} & AAAI'22 & 24.998 & 0.871 & 0.117 & 17.433 & 0.831 & 0.176 & 24.807 & 0.919 & 0.067 & 17.42 & 358.4 \\
SNR-Aware \cite{xu2022snr} & CVPR’22 & 24.612 & 0.842 & 0.152 & 21.480 & 0.849 & 0.163 & 24.140 & 0.928 & 0.056 & 4.01 & 26.35 \\
Bread \cite{guo2023low} & IJCV'23 & 25.299 & 0.847 & 0.155 & 20.830 & 0.847 & 0.174 & 17.630 & 0.919 & 0.091 & 2.02 & 19.85 \\
LLFormer \cite{wang2023ultra} & AAAI’23 & 25.758 & 0.823 & 0.167 & 20.056 & 0.792 & 0.211 & 24.038 & 0.909 & 0.066 & 24.55 & 22.52 \\
RetinexFormer \cite{cai2023retinexformer} & ICCV’23 & 27.140 & 0.850 & 0.129 & 22.794 & 0.840 & 0.171 & 25.670 & 0.930 & 0.059 & 1.53 & 15.85 \\
GSAD \cite{hou2023global} & NeurIPS’23 & 27.605 & 0.876 & 0.092 & 20.153 & 0.846 & 0.113 & 24.472 & 0.929 & 0.051 & 17.36 & 442.02 \\
UHDFormer \cite{wang2024correlation} & AAAI’24 & 26.388 & 0.856 & 0.136 & 19.71 & 0.832 & \textbf{\textcolor{red}{0.075}} & 24.480 & 0.927 & \textbf{\textcolor{red}{0.030}} & {{0.339}} & {{3.24}} \\
RetinexMamba \cite{bai2024retinexmamba} & ICONIP’25 & 26.343 & 0.838 & 0.136 & 22.453 & 0.844 & 0.174 & \textit{\textcolor{cyan}{25.887}} & 0.935 & 0.055 & 4.59 & 42.82 \\
HVI-CIDNet \cite{yan2025hvi} & CVPR’25 & \textit{\textcolor{cyan}{28.201}} & \textit{\textcolor{cyan}{0.889}} & \textit{\textcolor{cyan}{0.079}} & \textit{\textcolor{cyan}{24.111}} & \textit{\textcolor{cyan}{0.871}} & 0.108 & 25.705 & \textit{\textcolor{cyan}{0.942}} & 0.045 & 1.88 & 7.57 \\
\hline
\textbf{Ours} & \textbf{Ours} & \textbf{\textcolor{red}{28.560}} & \textbf{\textcolor{red}{0.893}} & \textbf{\textcolor{red}{0.068}} & \textbf{\textcolor{red}{24.822}} & \textbf{\textcolor{red}{0.882}} & \textit{\textcolor{cyan}{0.078}} & \textbf{\textcolor{red}{26.117}} & \textbf{\textcolor{red}{0.947}} & \textit{\textcolor{cyan}{0.039}} & 2.51 & 8.69 \\
\specialrule{1.2pt}{0pt}{0pt}
\end{tabular}
}
\end{table*}

\begin{figure*}[t]
    \centering
    \begin{subfigure}{\textwidth}
        \centering
        \includegraphics[width=0.12\textwidth]{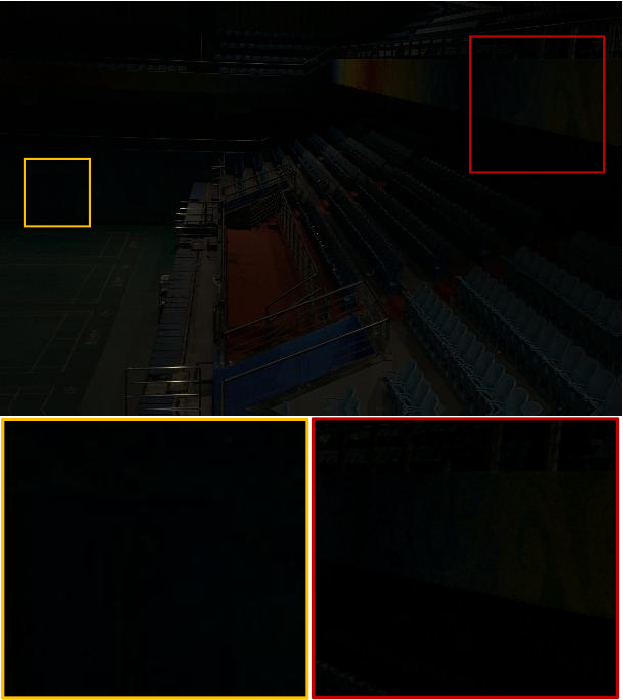}
        \includegraphics[width=0.12\textwidth]{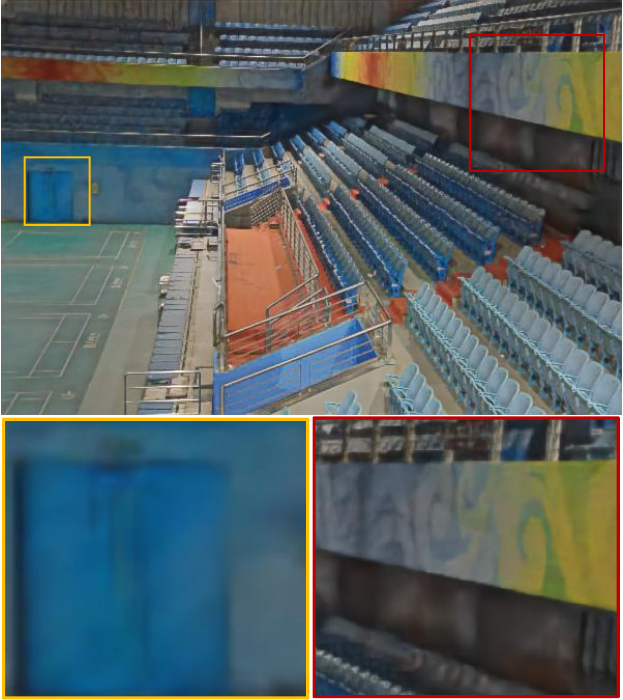}
        \includegraphics[width=0.12\textwidth]{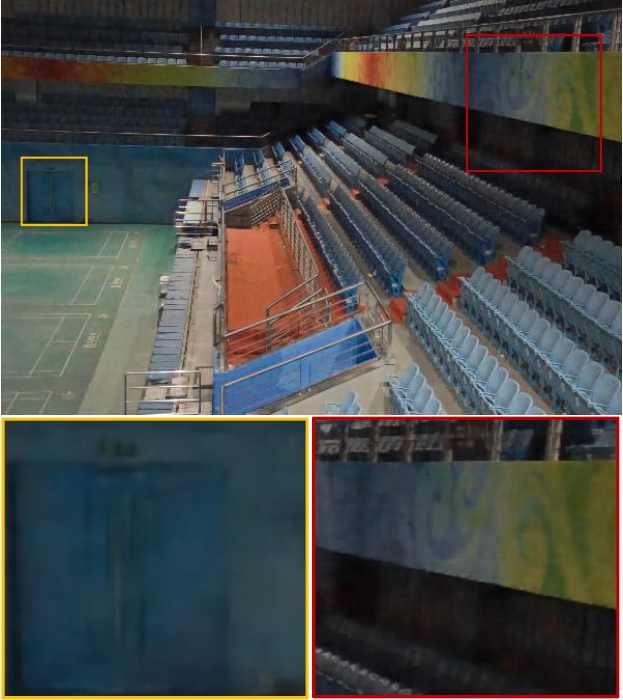}
        \includegraphics[width=0.12\textwidth]{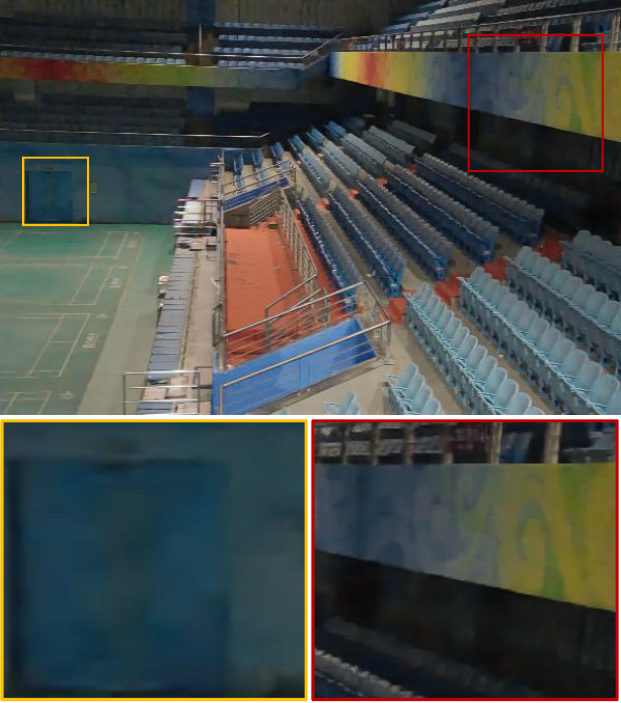}
        \includegraphics[width=0.12\textwidth]{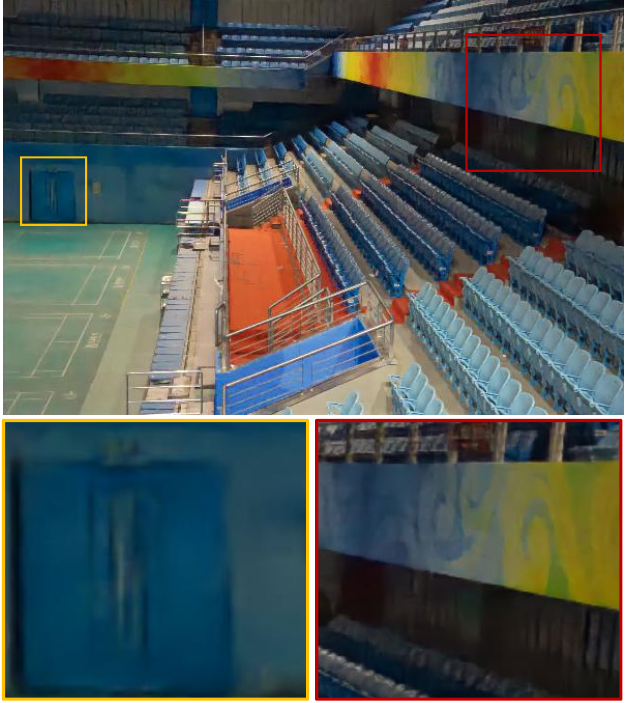}
        \includegraphics[width=0.12\textwidth]{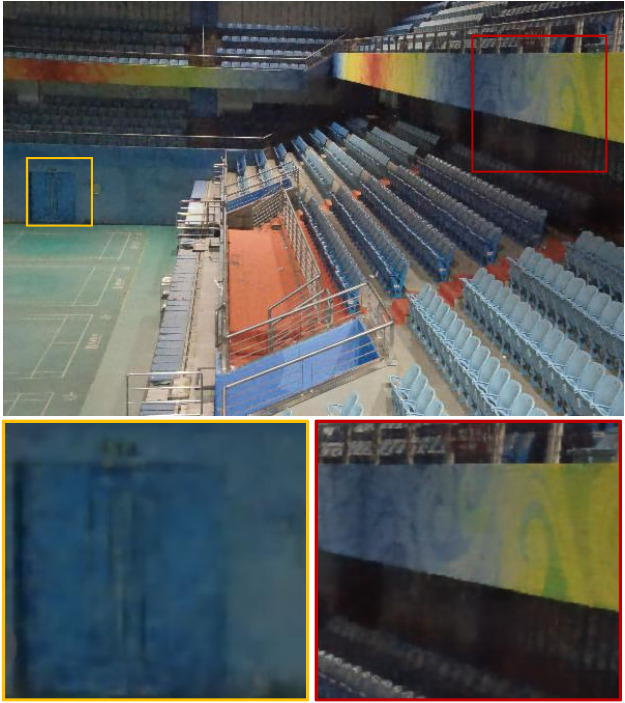}
        \includegraphics[width=0.12\textwidth]{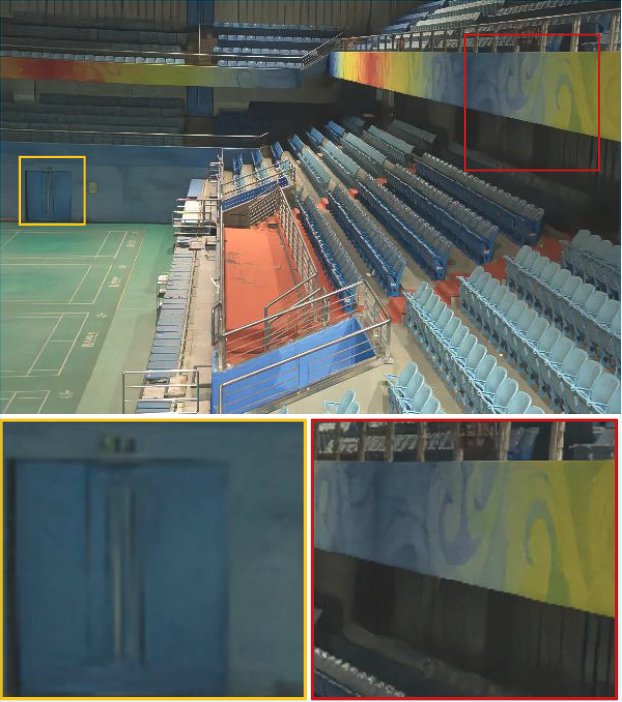}
        \includegraphics[width=0.12\textwidth]{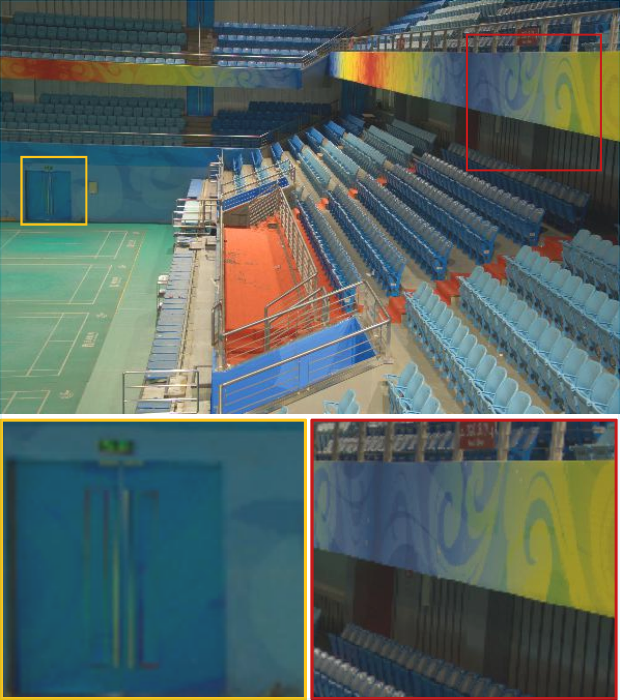}
    \end{subfigure}

    \vspace{0pt}
    \begin{subfigure}{\textwidth}
        \centering
        \includegraphics[width=0.12\textwidth]{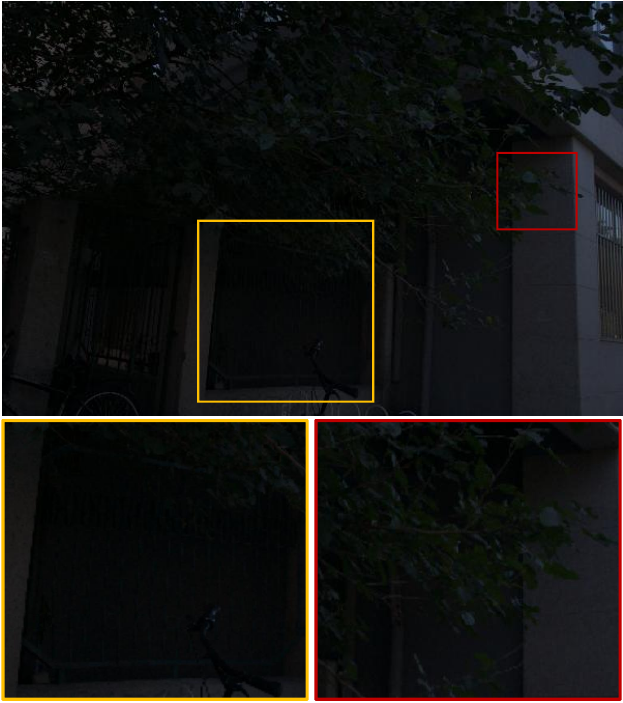}
        \includegraphics[width=0.12\textwidth]{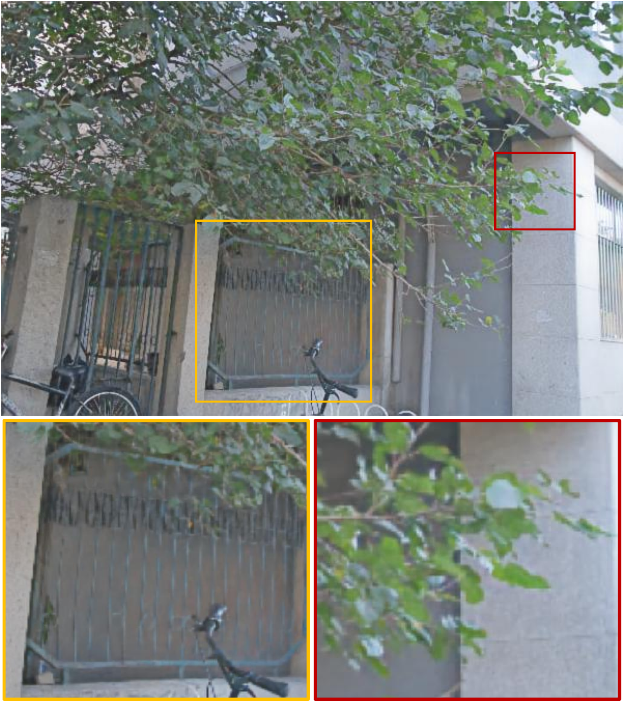}
        \includegraphics[width=0.12\textwidth]{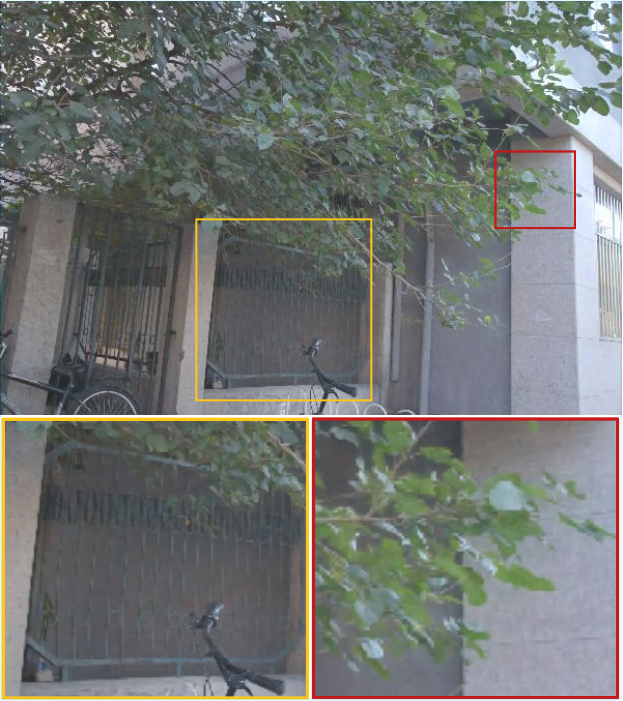}
        \includegraphics[width=0.12\textwidth]{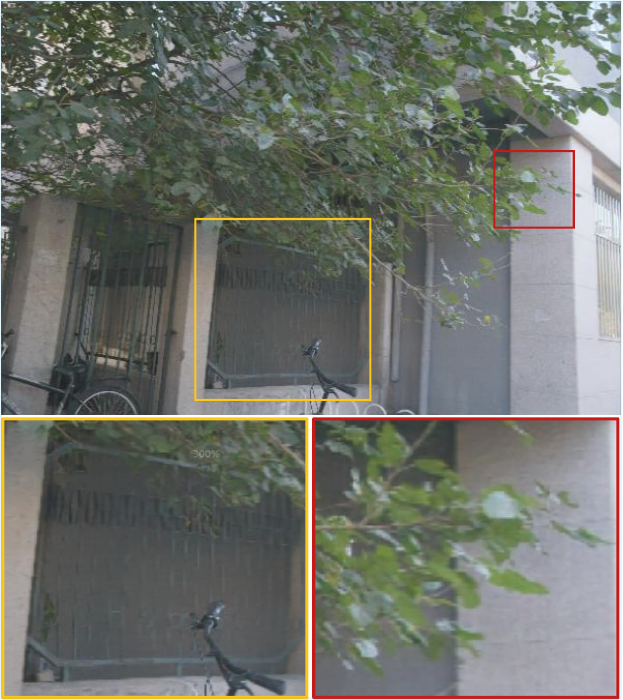}
        \includegraphics[width=0.12\textwidth]{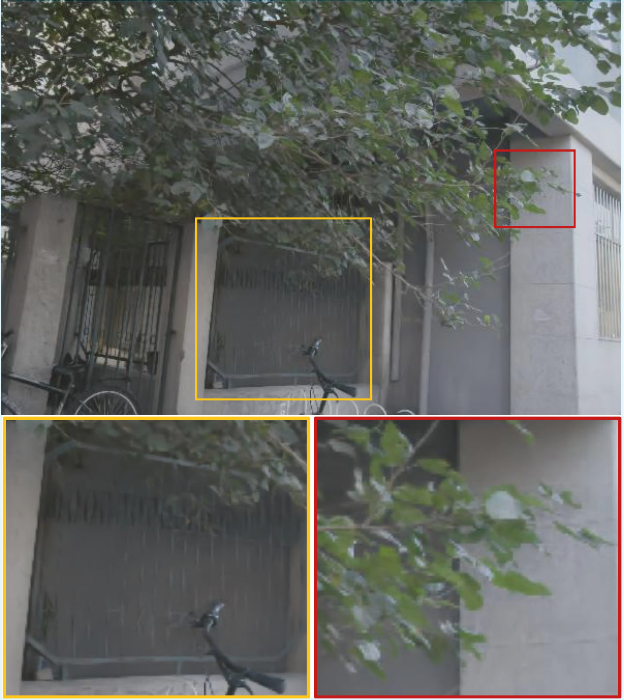}
        \includegraphics[width=0.12\textwidth]{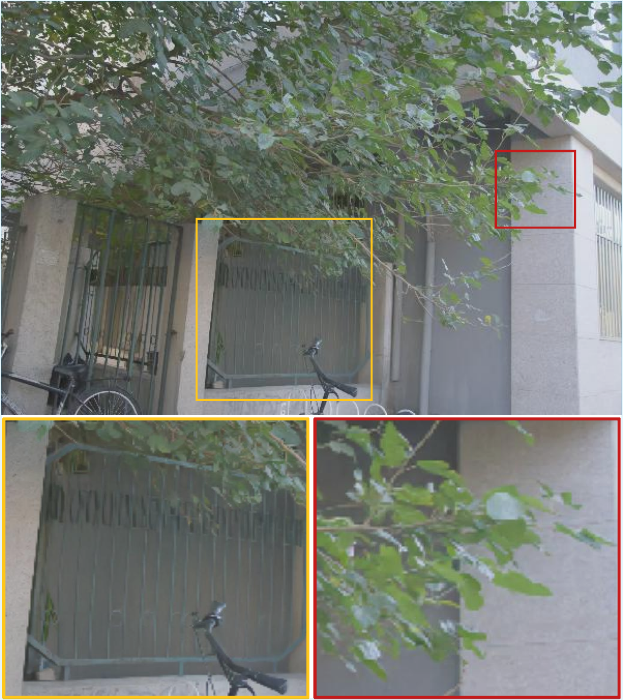}
        \includegraphics[width=0.12\textwidth]{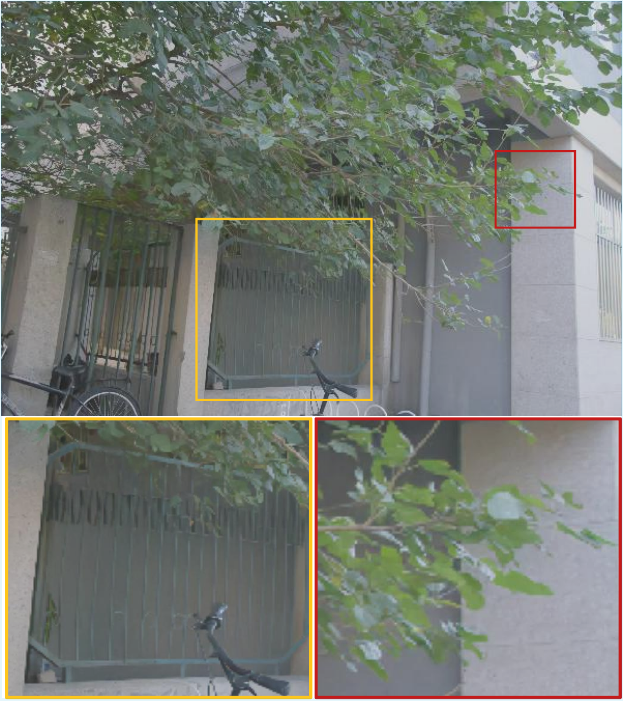}
        \includegraphics[width=0.12\textwidth]{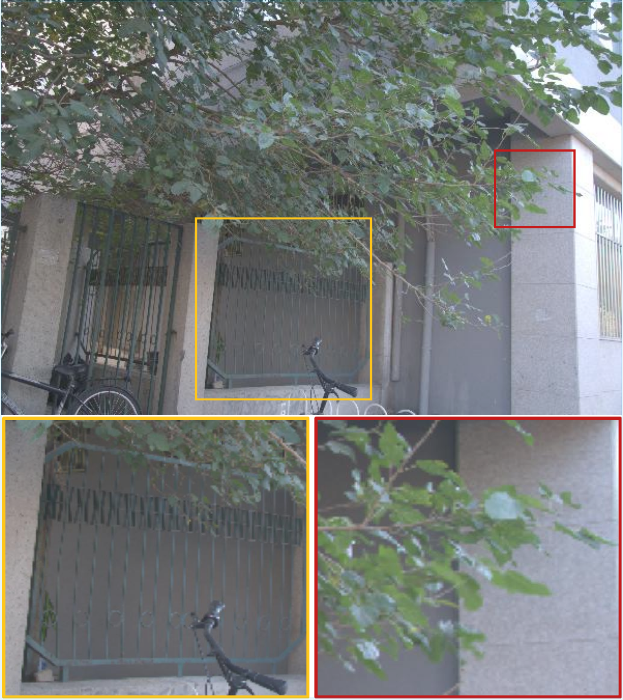}
    \end{subfigure}

    \vspace{0pt}
    \begin{subfigure}{\textwidth}
        \footnotesize
        \makebox[0.12\textwidth]{Input}%
        \makebox[0.15\textwidth]{KinD}%
        \makebox[0.11\textwidth]{MIRNet}%
        \makebox[0.12\textwidth]{SNR}%
        \makebox[0.13\textwidth]{CIDNet}%
        \makebox[0.12\textwidth]{UHDFormer}%
        \makebox[0.13\textwidth]{Ours}%
        \makebox[0.12\textwidth]{GT}%
    \end{subfigure}

    \caption{Visual comparison of the enhanced images yielded by different SOTA methods on LOLv1 (upper) and LOLv2 (lower).}
    \Description{A multi-panel figure comparing image enhancement results. The top row shows results on LOLv1 dataset, and the bottom row shows results on LOLv2. Columns represent Input, KinD, MIRNet, SNR, CIDNet, UHDFormer, Ours, and Ground Truth. Our method shows better detail retention and color balance.}
    \label{fig:comparison}
\end{figure*}
\setlength{\textfloatsep}{10pt}
\setlength{\floatsep}{10pt}
\setlength{\intextsep}{10pt}
\setlength{\abovecaptionskip}{10pt}
\setlength{\belowcaptionskip}{0pt}
\setlength{\textfloatsep}{6pt}
\setlength{\floatsep}{6pt}
\setlength{\intextsep}{6pt}
\setlength{\abovecaptionskip}{2pt}
\setlength{\belowcaptionskip}{4pt}

\begin{table}[t]
\centering

\caption{Quantitative comparison on SICE (SICE-Grad, SICE-Mix) and Sony-Total-Dark datasets. 
}
\label{tab:challenging_comparison}
\setlength{\tabcolsep}{6pt} 
\small

\resizebox{\columnwidth}{!}{%
\begin{tabular}{@{\hskip0pt} c | c c | c c | c c @{\hskip0pt}}
\specialrule{1.2pt}{0pt}{0pt} 

\multirow{2}{*}{\centering\arraybackslash\textbf{Methods}} &
\multicolumn{2}{c|}{\textbf{SICE-Grad}} &
\multicolumn{2}{c|}{\textbf{SICE-Mix}} &
\multicolumn{2}{c}{\textbf{Sony-Total-Dark}} \\

\cline{2-7}
 & PSNR$\uparrow$ & SSIM$\uparrow$ & PSNR$\uparrow$ & SSIM$\uparrow$ & PSNR$\uparrow$ & SSIM$\uparrow$ \\
\hline

RetinexNet \cite{wei2018deep} & 12.450 & 0.619 & 12.397 & 0.606 & 15.695 & 0.395 \\
Zero-DCE \cite{guo2020zero} & 12.475 & 0.644 & 12.428 & 0.633 & 14.087 & 0.090 \\
RUAS \cite{liu2021retinex} & 8.628 & 0.494 & 8.684 & 0.493 & 12.622 & 0.081 \\
URetinexNet \cite{wu2022uretinex} & 10.894 & 0.610 & 10.903 & 0.600 & 12.622 & 0.081 \\
LLFlow \cite{wang2022low} & 12.737 & 0.617 & 12.737 & 0.617 & 16.226 & 0.367 \\
LEDNet \cite{wang2019lednet} & 12.551 & 0.576 & 12.668 & 0.579 & 20.830 & 0.648 \\
HVI-CIDNet \cite{yan2025hvi} & \textit{\textcolor{cyan}{13.446}} & \textit{\textcolor{cyan}{0.648}} & \textit{\textcolor{cyan}{13.425}} & \textit{\textcolor{cyan}{0.636}} & \textit{\textcolor{cyan}{22.904}} & \textit{\textcolor{cyan}{0.676}} \\
\hline
\textbf{Ours} & \textbf{\textcolor{red}{13.901}} & \textbf{\textcolor{red}{0.665}} & \textbf{\textcolor{red}{13.848}} & \textbf{\textcolor{red}{0.653}} & \textbf{\textcolor{red}{23.275}} & \textbf{\textcolor{red}{0.684}} \\
\specialrule{1.2pt}{0pt}{0pt} 
\end{tabular}
} 
\end{table}
\begin{table}[t]
\centering
\caption{Quantitative comparison (NIQE$\downarrow$) on five real-world unpaired datasets (DICM, LIME, MEF, NPE, and VV).  
}
\label{tab:unpaired_niqe}
\setlength{\tabcolsep}{6pt} 
\renewcommand{\arraystretch}{0.93}

\resizebox{\columnwidth}{!}{%
\footnotesize
\begin{tabular}{@{\hskip0pt} c | c | c | c | c | c @{\hskip0pt}}
\specialrule{1.0pt}{0pt}{0pt} 

\textbf{Methods} & \textbf{DICM}  & \textbf{LIME}  & \textbf{MEF}  & \textbf{NPE}   & \textbf{VV\quad}   \\
\hline

Zero-DCE \cite{guo2020zero} & 4.580 & 5.820 & 4.930 & 4.530 &{4.810\quad} \\
KinD \cite{zhang2019kindling} & 3.614 & 4.772 & 4.819 & 4.175 &{3.835\quad}

\\
FECNet \cite{huang2022deep} & 4.139 & 6.041 & 4.707 & 4.500 &{3.346\quad} \\
MIRNet \cite{zamir2020learning} & 4.042 & 6.453 & 5.504 & 5.235 &{4.735\quad} \\
SMG \cite{xu2023low} & 4.733 & 5.451 & 5.754 & 5.208 &{4.884\quad} \\
Bread \cite{guo2023low} & 4.179 & 4.717 & 5.369 & 4.160 &{3.304\quad} \\
FourLLIE \cite{wang2023fourllie} & \textcolor{red}{\textbf{3.374}} & 4.402 & 4.362 & 3.909 & \textcolor{cyan}{\textit{3.168\quad}} \\
HVI-CIDNet \cite{yan2025hvi} & 3.750 & \textcolor{cyan}{\textit{4.151}} & \textcolor{cyan}{\textit{3.572}} & \textcolor{red}{\textbf{3.746}} & {3.212\quad} \\
\hline
\textbf{Ours} & \textcolor{cyan}{\textit{3.505}} & \textcolor{red}{\textbf{4.022}} & \textcolor{red}{\textbf{3.522}} & \textcolor{cyan}{\textit{3.753}} & \textcolor{red}{\textbf{3.162\quad}} \\

\specialrule{1.0pt}{0pt}{0pt} 
\end{tabular}
} 
\end{table}
\setlength{\textfloatsep}{10pt}
\setlength{\floatsep}{10pt}
\setlength{\intextsep}{10pt}
\setlength{\abovecaptionskip}{10pt}
\setlength{\belowcaptionskip}{0pt}

Compared with the baseline method HVI-CIDNet~\cite{yan2025hvi}, our approach achieves consistent and significant improvements across all metrics. Notably, the LPIPS scores are substantially reduced on all three subsets (by 13.9\%, 27.8\%, and 13.3\%, respectively). This strongly demonstrates that the RHVI-FDD effectively captures and responds to the underlying physical mechanisms of low-light degradation, transcending the pixel-level fidelity pursued by conventional black-box deep learning methods~\cite{li2021low}. Our advantage is most evident on the Sony-Total-Dark dataset, captured under extremely low illumination with heavy sensor noise. The Max-RGB-based illumination estimation in the original HVI is highly noise-sensitive, while our IRM robustly refines the illumination map. This targeted design leads to superior performance on SID, validating the effectiveness and robustness of RHVI transform in ultra-low-light conditions.

To further assess generalization, we test the module trained on LOLv2-Real dataset across five unpaired real-world datasets. As shown in Table~\ref{tab:unpaired_niqe}, our method demonstrates outstanding generalization, achieving the best or second-best 
NIQE scores on all five datasets. This highlights the model's superior capability to enhance low-light images across diverse real-world conditions.

\begin{table*}[t]
\centering
\setlength{\abovecaptionskip}{0pt}
\setlength{\belowcaptionskip}{3pt}
\caption{Results of applying RHVI-FDD as a plug-in to various LLIE methods on the LOLv2-Real dataset. Values in brackets represent the absolute performance gain after integration. \textcolor{red}{Red} values indicate performance improvement (corresponding to higher PSNR/SSIM$\uparrow$ or lower LPIPS$\downarrow$), while \textcolor{green}{green} values indicate a performance drop. Our RHVI-FDD has 0.63M parameters.}
\label{tab:plug_and_play}
\setlength{\tabcolsep}{3pt}
\small

\resizebox{\textwidth}{!}{
\begin{tabular}{@{\hskip2pt} c | c c c c c c c @{\hskip2pt}}
\specialrule{1.2pt}{0pt}{0pt}
\rowcolor{gray!15}
\textbf{Methods} & \textbf{RetinexNet}~\cite{wei2018deep} & \textbf{LEDNet}~\cite{wang2019lednet} & \textbf{SNR-Aware}~\cite{xu2022snr} & \textbf{LLFormer}~\cite{wang2023ultra} & \textbf{LCDBNet}~\cite{wang2024division} & \textbf{GSAD}~\cite{hou2023global} & \textbf{CIDNet}~\cite{yan2025hvi}  \\
\hline
PSNR $\uparrow$ 
& 20.803 (\textcolor{red}{+4.789}) 
& 23.307 (\textcolor{red}{+3.529}) 
& 22.017 (\textcolor{red}{+0.697}) 
& 22.710 (\textcolor{red}{+2.678}) 
& 24.029 (\textcolor{red}{+3.682}) 
& 23.821 (\textcolor{red}{+3.668}) 
& 24.822 (\textcolor{red}{+0.711}) \\
SSIM $\uparrow$ 
& 0.652 (\textcolor{red}{+0.108}) 
& 0.838 (\textcolor{red}{+0.031}) 
& 0.839 (\textcolor{green}{-0.006}) 
& 0.866 (\textcolor{red}{+0.074}) 
& 0.833 (\textcolor{red}{+0.053}) 
& 0.874 (\textcolor{red}{+0.029}) 
& 0.882 (\textcolor{red}{+0.011}) \\
LPIPS $\downarrow$ 
& 0.351 (\textcolor{red}{-0.106}) 
& 0.113 (\textcolor{red}{-0.013}) 
& 0.111 (\textcolor{red}{-0.067}) 
& 0.121 (\textcolor{red}{-0.090}) 
& 0.126 (\textcolor{red}{-0.048}) 
& 0.101 (\textcolor{red}{-0.012}) 
& 0.078 (\textcolor{red}{-0.030}) \\
\hline
Params / M 
& 0.84 & 7.4 & 4.01 & 24.55 & 7.36 & 17.36 & 1.88 \\

Model Type 
& CNN & CNN & Transformer & Transformer & CNN & Diffusion & Transformer \\

Branch 
& Dual & Single & Single & Single & Dual & Single & Dual \\
\specialrule{1.2pt}{0pt}{0pt}
\end{tabular}
}
\end{table*}

\begin{table}[t]
\centering
\setlength{\abovecaptionskip}{0pt}
\setlength{\belowcaptionskip}{3pt}
\caption{Ablation study on different components of our framework on LOLv2-Real.}
\label{tab:ablation_final}

\setlength{\tabcolsep}{6pt}
\renewcommand{\arraystretch}{1.2}

\resizebox{\columnwidth}{!}{%
\footnotesize
\begin{tabular}{l | c c c}
\specialrule{1.2pt}{0pt}{0pt}
\rowcolor{gray!15}
\textbf{Methods} & \textbf{PSNR $\uparrow$} & \textbf{SSIM $\uparrow$} & \textbf{LPIPS $\downarrow$} \\
\hline

Ours (RHVI $\to$ HSV)        & 21.698 & 0.822 & 0.169 \\
Ours (RHVI $\to$ HVI)        & 23.921 & 0.867 & 0.120 \\
Ours w/o IRM                 & 24.326 & 0.871 & 0.115 \\

\hline
Ours w/o GCM (Low-freq)      & 23.952 & 0.865 & 0.110 \\
Ours w/o DRG (Mid-freq)      & 23.518 & 0.852 & 0.132 \\
Ours w/o ANSU (High-freq)    & 24.215 & 0.873 & 0.098 \\
Ours w/o ACGF (Gated)        & 24.120 & 0.869 & 0.105 \\
Ours w/o FDD                 & 23.864 & 0.863 & 0.125 \\

\hline
\textbf{Full Model}          & \textbf{24.822} & \textbf{0.882} & \textbf{0.078} \\
\specialrule{1.2pt}{0pt}{0pt}
\end{tabular}
}
\end{table}
\subsection{Qualitative Comparison}
To visually demonstrate the superiority of our approach, Fig.~\ref{fig:comparison} presents qualitative comparisons across various challenging scenes. Our method generates more natural brightness, finer detail restoration, and fewer artifacts, even in regions with complex textures.

For extremely dark scenes with severe noise, as shown in the second row of Fig.~\ref{fig:comparison} from the SID dataset, the baseline's reliance on the Max-RGB theory introduces severe color artifacts in dark areas. In contrast, our RHVI suppresses these noise-induced artifacts at the source, producing visually consistent and artifact-free results.
Notably, the advantage of our method is particularly evident in complex scenes with intertwined noise and details. Existing methods suffer from a trade-off between denoising and detail preservation---either over-smoothing fine textures or amplifying noise in flat areas~\cite{tian2023survey}, while our method excels due to the FDD.
\subsection{Ablation Studies}
We conducted a series of ablation experiments by progressively removing key components from our framework, with quantitative results summarized in Table~\ref{tab:ablation_final}. As shown, the complete model consistently achieves the best performance across all metrics. 

At the macro level, the performance drops in variants using traditional HSV~\cite{sural2002segmentation} or standard HVI confirm that our RHVI transform effectively resolves noise-sensitivity issues. Furthermore, removing the IRM or auxiliary supervision leads to notable declines, highlighting their importance in suppressing sensor noise and guiding robust illumination estimation. At the micro level, excluding any individual frequency expert or the ACGF gating mechanism results in clear performance degradation. Overall, the substantial drop without FDD underscores its role as the core driver of our method.
\subsection{Generalizability Analysis}
Notably, the RHVI-FDD does not rely on specific dual-branch host networks, but is broadly compatible with mainstream end-to-end feature enhancement methods. We evaluate the generalizability of RHVI-FDD by integrating it as a plug-and-play module into representative benchmarks (Table~\ref{tab:plug_and_play}). Specifically, RHVI is incorporated via input tensor reconstruction by concatenating decoupled components into a 3-channel representation, while FDD is inserted into the bottleneck of single-stream models or the enhancement branch of dual-stream architectures. 

\setlength{\textfloatsep}{6pt}
\setlength{\floatsep}{6pt}
\setlength{\intextsep}{6pt}
\setlength{\abovecaptionskip}{2pt}
\setlength{\belowcaptionskip}{0pt}
\begin{figure}[t]
    \centering

    \begin{subfigure}{\linewidth}
        \centering
        \includegraphics[width=0.32\linewidth]{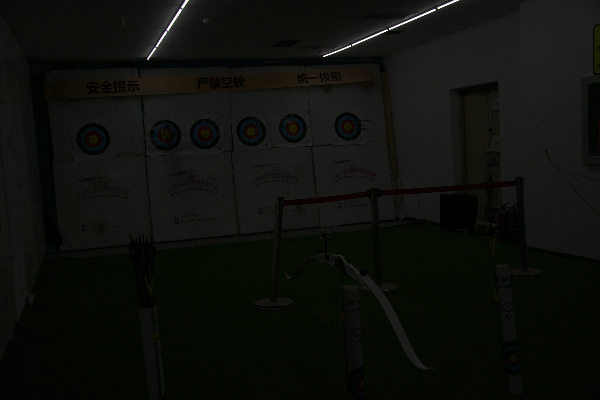}
        \includegraphics[width=0.32\linewidth]{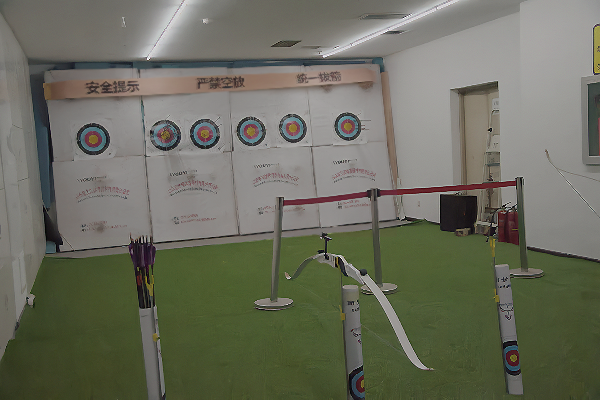}
        \includegraphics[width=0.32\linewidth]{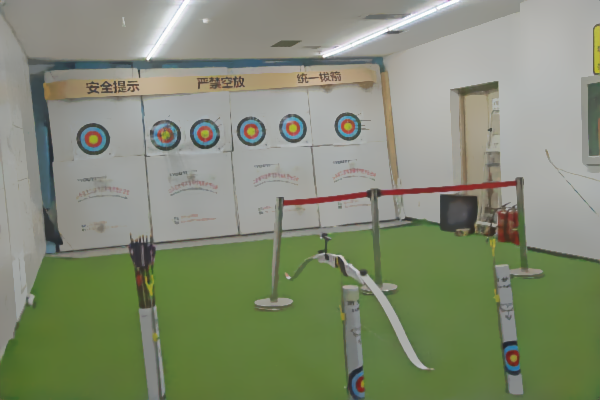}
    \end{subfigure}

    \vspace{-3pt} 
    \begin{subfigure}{\linewidth}
        \centering
        \footnotesize
        \parbox{0.32\linewidth}{\centering (a) Input}%
        \parbox{0.32\linewidth}{\centering (b) Ours w/o L}%
        \parbox{0.32\linewidth}{\centering (c) Ours w/o M}%
    \end{subfigure}

    \vspace{5pt} 
    \begin{subfigure}{\linewidth}
        \centering
        \includegraphics[width=0.32\linewidth]{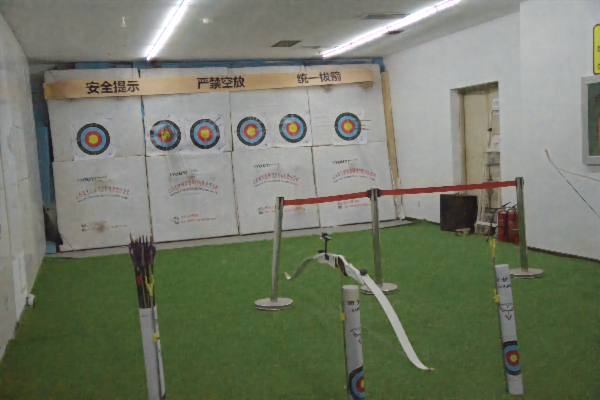}
        \includegraphics[width=0.32\linewidth]{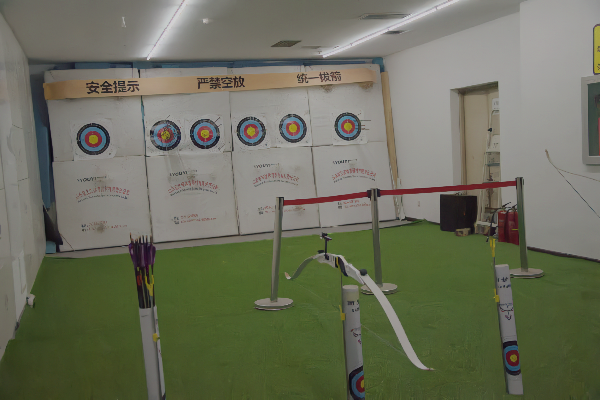}
        \includegraphics[width=0.32\linewidth]{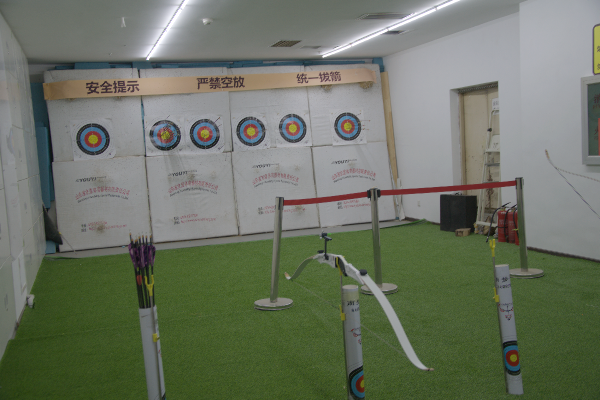}
    \end{subfigure}

    \vspace{-3pt} 
    \begin{subfigure}{\linewidth}
        \centering
        \footnotesize
        \parbox{0.32\linewidth}{\centering (d) Ours w/o H}%
        \parbox{0.32\linewidth}{\centering (e) Full}%
        \parbox{0.32\linewidth}{\centering (f) GT}%
    \end{subfigure}
    \Description{A grid comparison showing the ablation study results of our hierarchical decoupling paradigm. 
(a) is the underexposed and noisy input. 
(b) without the low-frequency component shows insufficient brightness and poor global tone. 
(c) without the mid-frequency component exhibits blurred structures and a loss of fine textures. 
(d) without the high-frequency component shows significant residual noise and artifacts. 
(e) is the full model, which achieves natural illumination and sharp details comparable to the ground truth (f).}
    \caption{Ablation study on the contributions of low, mid, and high-frequency components.}
    \label{fig:abla_study}
\end{figure}
\setlength{\textfloatsep}{10pt}
\setlength{\floatsep}{10pt}
\setlength{\intextsep}{10pt}
\setlength{\abovecaptionskip}{10pt}
\setlength{\belowcaptionskip}{0pt}
As shown in Table~\ref{tab:plug_and_play}, RHVI-FDD consistently improves performance across diverse architectures, yielding PSNR improvements while simultaneously achieving substantial LPIPS reductions. With only 0.63M additional parameters, RHVI-FDD offers a lightweight and physically interpretable feature decoupling mechanism, enabling strong generalization across LLIE models.

\section{Conclusion}
In this paper, we systematically model and analyze the intrinsic problems of low-light images, which present two hierarchical challenges: luminance-chrominance coupling and noise-detail entanglement. Correspondingly, we propose a novel hierarchical decoupling framework (RHVI-FDD), which achieves robust luminance-chrominance decoupling at the macro level and leverages a frequency-domain divide-and-conquer strategy to decouple feature entanglement at the micro level. Extensive experiments validate the superiority, effectiveness, and generalization capability of our method. Furthermore, we envision that the proposed hierarchical and frequency-based decoupling strategy will inspire new perspectives for broader image restoration tasks.


\bibliographystyle{ACM-Reference-Format}

\bibliography{main}


\end{document}